%% file: main.tex
\documentclass{article}
\usepackage{ijcai26}

\usepackage{times}
\usepackage{soul}
\usepackage{url}
\usepackage[hidelinks]{hyperref}
\usepackage[utf8]{inputenc}
\usepackage[small]{caption}
\usepackage{graphicx}
\usepackage{booktabs}
\usepackage{algorithm}
\usepackage{algorithmic}
\usepackage[switch]{lineno}
\usepackage{subcaption}

\usepackage{xparse}
\usepackage{xstring}

\usepackage[table]{xcolor}                 
\definecolor{ease_darkblue}{HTML}{144F78}
\colorlet{background}{ease_darkblue!8} 
\colorlet{foreground}{ease_darkblue!50!white} 
\colorlet{ease_darkblueblue}{ease_darkblue!75!blue}
\colorlet{foregroundblue}{foreground!75!blue}
\definecolor{ease_darkgreen}{HTML}{62A27A}
\definecolor{ease_lightgreen}{HTML}{CDE5C1}
\colorlet{background2}{ease_lightgreen}
\colorlet{ease_darkgreengreen}{ease_darkgreen!50!green}
\colorlet{foreground2}{ease_darkgreengreen!55!black}
\definecolor{ease_darkorange}{HTML}{E29C47}
\definecolor{ease_lightorange}{HTML}{F2E6BD}
\colorlet{foreground3}{ease_darkorange}
\colorlet{background3}{ease_lightorange}
\colorlet{ease_darkorangeorange}{ease_darkorange!75!orange}
\definecolor{ease_darkred}{HTML}{A74D4B}
\definecolor{ease_lightred}{HTML}{EFDACE}
\colorlet{foreground4}{ease_darkred}
\colorlet{background4}{ease_lightred}
\colorlet{gls_links}{gray!50}

\definecolor{myyellow1}{rgb}{1.0, 0.85, 0.5} 
\definecolor{myyellow2}{rgb}{1.0, 0.7, 0.0} 
\definecolor{myyellow3}{rgb}{1.0, 0.6, 0.0} 
\definecolor{myyellow4}{rgb}{0.6, 0.41, 0.0} 

\definecolor{myred1}{rgb}{1.0, 0.65, 0.5} 
\definecolor{myred2}{rgb}{1.0, 0.31, 0.0} 
\definecolor{myred3}{rgb}{1.0, 0.25, 0.0} 
\definecolor{myred4}{rgb}{0.6, 0.188, 0.0} 

\definecolor{falsered}{HTML}{FF5024}
\definecolor{unknowngray}{HTML}{8F959E}
\definecolor{truegreen}{HTML}{B6E5A0}

\definecolor{lcfsmNotStartedColor}{HTML}{8F959E} 
\definecolor{lcfsmRunningColor}{HTML}{28A745} 
\definecolor{lcfsmPausedColor}{HTML}{E6AC00} 
\definecolor{lcfsmStoppedColor}{HTML}{DC3545} 

\definecolor{mygreen1}{rgb}{0.7, 1.0, 0.7}
\definecolor{mygreen4}{rgb}{0.0, 0.4, 0.0} 

\definecolor{myblue1}{rgb}{0.7, 0.85, 1.0}
\definecolor{myblue4}{rgb}{0.0, 0.2, 0.6}

\definecolor{lightred}{rgb}{1.0, 0.7, 0.7}
\definecolor{darkred}{rgb}{0.6, 0.0, 0.0}

\definecolor{mySTDpurple}{rgb}{0.588, 0.490, 0.784}
\definecolor{mySTDblue}{rgb}{0.0, 0.588, 0.784}
\definecolor{mySTDyellow}{rgb}{0.784, 0.588, 0.0}

\definecolor{darkgrey}{HTML}{575757}
\definecolor{lightgrey}{HTML}{E0E0E0} 
\definecolor{lightgreen}{HTML}{EDFFFC}
\definecolor{greenerlightgreen}{HTML}{C9FBD4}
\definecolor{darkgreen}{HTML}{1C8575}
\definecolor{midgreen}{HTML}{00C4A7}
\definecolor{darkyellowaccent}{HTML}{C4A605}
\definecolor{lightyellowaccent}{HTML}{FCE882}
\definecolor{darkredaccent}{HTML}{851C3F}
\definecolor{lightredaccent}{HTML}{DB5784}
\definecolor{eyecancergreen}{HTML}{7EFFEC}

\colorlet{background}{lightgrey}
\colorlet{foregroundros}{myyellow4}
\colorlet{foregroundlib}{myred4}
\colorlet{foreground2}{darkgrey}

\usepackage{amsmath,amssymb,amsthm,mathtools}

\usepackage[most]{tcolorbox}
\tcbuselibrary{theorems}

\tcbset{
  theorem style/.style={
    enhanced,
    colback=white,
    colframe=black!70,
    rounded corners,
    boxrule=0.7pt,
    arc=3pt,
    left=6pt,
    right=6pt,
    top=6pt,
    bottom=6pt,
    fonttitle=\bfseries,
    colbacktitle=white,
    coltitle=black,
    attach boxed title to top left={yshift=-2mm,xshift=2mm},
    boxed title style={
      colframe=white,
      sharp corners,
      boxrule=0pt,
      top=0pt,
      bottom=0pt,
      left=0pt,
      right=0pt
    }
  },
}

\newtcbtheorem[number within=section]{definition}{Definition}%
{theorem style}{def}

\newtcbtheorem[number within=section]{proposition}{Proposition}%
{theorem style}{prop}

\newtcbtheorem[number within=section]{remark}{Remark}%
{theorem style}{rem}

\usepackage{enumitem}
\usepackage{tikz}
\usepackage[nolist]{acronym}
\setlist{nosep,topsep=2pt,partopsep=0pt,leftmargin=*,itemsep=1pt,parsep=0pt}

\title{Closing the Motion Execution Gap:\\From Semantic Motion Task Constraints to Kinematic Control}

\author{
Simon Stelter
\and
Vanessa Hassouna\and
Malte Huerkamp\And
Michael Beetz\\
\affiliations
University of Bremen\\
\emails
\{stelter, hassouna, huerkamp, mbeetz\}@uni-bremen.de
}
\input{cmds/names}

\input{cmds/tikz}
\input{cmds/general}

\input{cmds/transforms}

\input{cmds/world}

\input{cmds/monitors}
\input{cmds/qp}

\input{acronyms}

\begin{document}

\maketitle

\begin{abstract}
This paper addresses the Motion Execution Gap, the disconnect between high-level symbolic task descriptions using semantic constraints and executable robot motions.
\textbf{Motion Statecharts} are introduced as an executable symbolic representation for complex motions.
They allow the arbitrary arrangement of motion constraints, monitors or nested statecharts in parallel and sequence.
World-centric motion specification and generalization across embodiments are enabled through the use of a unified differentiable kinematic world model of both, robots and environments.
Motion execution is realized through a \ac{lmpc}-based implementation of the task-function approach, in which smooth transitions during task switches are ensured using jerk bounds.
Cross-platform transferability was demonstrated by deploying the method on eight robot platforms, operating in diverse environments.
The proposed framework is called Giskard and is available open source (\url{https://github.com/cram2/cognitive_robot_abstract_machine}).
\end{abstract}

\section{Introduction}

\begin{figure}
    \includegraphics[width=\columnwidth]{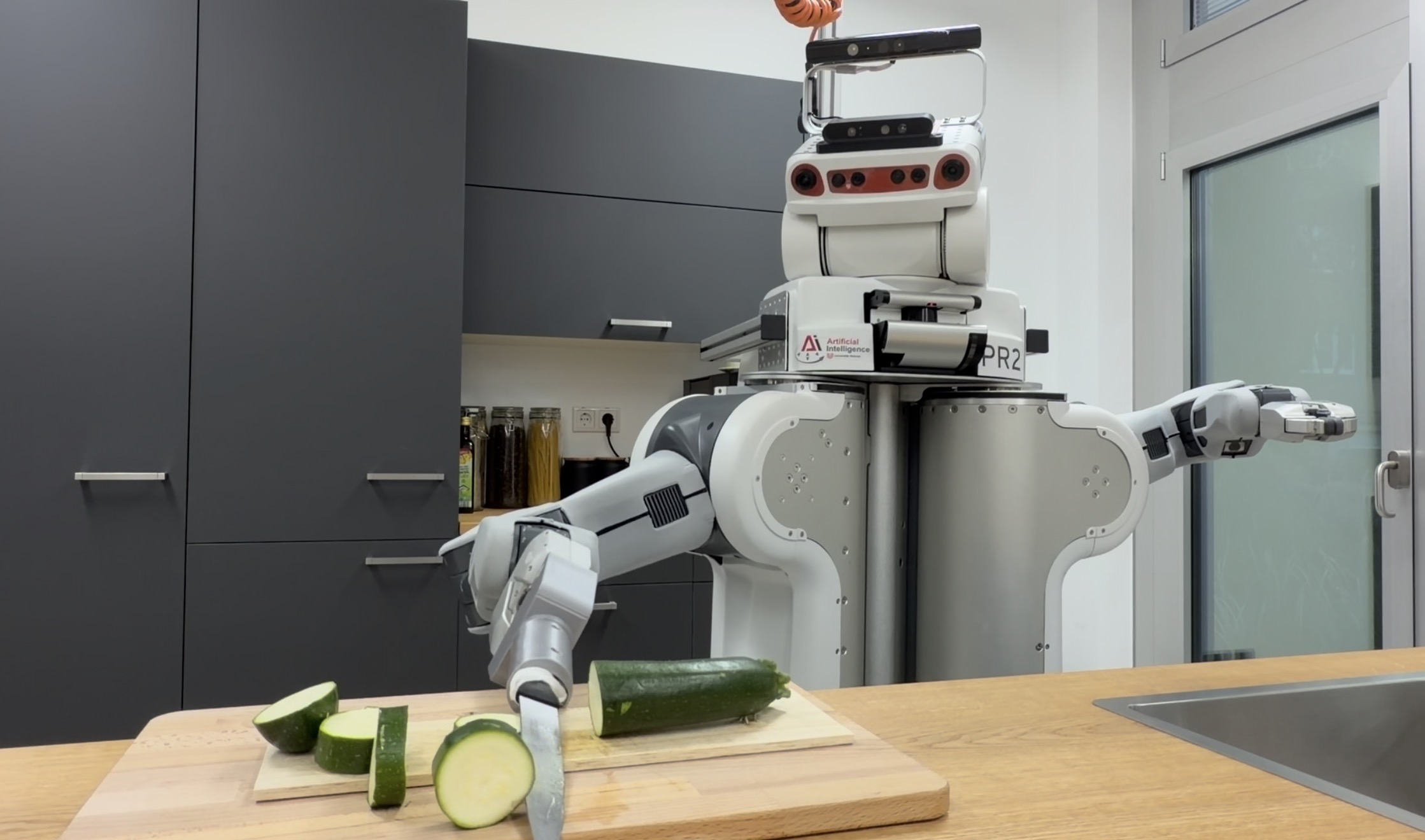}
    \caption{Illustrating the Motion Execution Gap: PR2 must generate constraint-satisfying trajectories from semantic instruction ``cut zucchini into small slices''.}
    \label{fig:cut_real}
\end{figure}

When describing robot motions, semantic constraints provide a natural and expressive approach. Consider ``make a salad'', as shown in Fig.~\ref{fig:cut_real} where a PR2 robot prepares ingredients by cutting a zucchini.
Such instructions are represented in cognitive robotics as symbolic tasks specifying \textbf{what} to achieve, not \textbf{how} to move.
High-level actions are decomposed into sub-tasks (cutting, grasping, placing) using Generative AI~\cite{gupta2024action} or plan executives~\cite{cram,pddl2}, often leveraging knowledge graphs~\cite{kuempel2025cookingactions}.
For example, ``cut the zucchini into small slices'' yields explicit constraints (slice thickness) and implicit ones (safe knife handling near humans), separating valid from invalid trajectories.
A system remains essential to convert these symbolic, constraint-rich descriptions into continuous motion satisfying all constraints.

Complete cognitive robot architectures such as SkiROS2~\cite{mayr2023skiros2}, ArmarX~\cite{wachter2016armarx}, and CRAM~\cite{beetz2023cram} are designed to achieve transferability across robots by hierarchically resolving symbolic motion tasks down to primitive Cartesian pose or joint-space goals.
Those primitive poses are typically further divided into mobile base and manipulator tasks~\cite{sandakalum2022motion} and they assume that tasks can be grounded at planning time~\cite{pan2024task}.
This design inherently limits the class of symbolic motion tasks that can be defined.
A task such as ``open the fridge'' cannot easily be reduced to these primitives, as it requires a discrete arch trajectory for the hand and the robot's mobile base to be positioned appropriately~\cite{bozcuouglu2018exchange}.
Motions requiring degrees of freedom, like pouring from a jug at an arbitrary angle, are more naturally expressed with geometric constraints~\cite{featurefunctions}.
This limitation constitutes the ``Motion Execution Gap''.

Addressing the motion execution gap through generative AI remains infeasible.
Due to their probabilistic nature, such methods cannot provide formal guarantees for constraint adherence~\cite{huang2025sad}.
Even state-of-the-art \acp{vla} require embodiment-specific post-training for transferability across robotic platforms~\cite{kim2025fine}.

A popular alternative is to formulate motions as constraint optimization problems.
The task function approach expressed as a \ac{qp}, often termed \ac{qp}-control~\cite{etasl,bouyarmane2018quadratic,giskard,sot}, is a promising variation.
Several works integrate \ac{qp}-based task-function control with symbolic motion representations:
In~\cite{pane2021autonomous}, a \ac{qp}-controller pairs with a \ac{fsm} representation, executing motion tasks in parallel while \ac{fsm} transitions serve as monitors.
~\cite{bolotnikova2021task} similarly combine \acp{fsm} with \ac{qp}-control for sequential/parallel multi-limb tasks.
~\cite{dominguez2022stack} attaches a low-frequency \ac{bt} to a high-frequency \ac{qp}-controller, dynamically adding/removing tasks via implicit end conditions for smoothness.
~\cite{diffbrum} employs \acp{bt} for pick-and-place sequencing, focusing on discrete motions rather than continuous task composition.
~\cite{rovida2018motion} extends \acp{bt} with parallel composites for concurrent impedance-controlled ``skills'', merging commands from parallel nodes.
\acp{fsm} and \acp{bt} are common choices, though \acp{fsm} suffer state explosion and \acp{bt} face computational scaling issues.

\acp{msc} are introduced as statecharts specialized for motion composition.
Statecharts, first introduced by \cite{harel1987statecharts}, have become a general term for extensions of \acp{fsm} aimed at alleviating the state explosion problem for specific use cases.
\acp{msc} enable symbolic reasoning over motion tasks and support online monitoring and adaptation akin to hybrid symbolic/neural planning.

However, task transitions commonly introduce discontinuities, typically mitigated by slowing down during transition~\cite{dominguez2022stack} or gradual task weight scaling~\cite{pane2021autonomous}, both limiting the motion representation.
For instance, single-task switches necessitate system-wide deceleration, impacting parallel tasks.
The proposed system instead formulates task functions as short-horizon \ac{lmpc} with explicit jerk limits, ensuring a smooth velocity motion profile.
\acp{msc} are thus allowed to swap motion tasks at any time without concern for the underlying controller.

In most of the aforementioned task function-based systems, only the robot kinematics are modeled.
In this paper, it is instead proposed to model the kinematics of the environment as well to enable motions to be defined in a world-centric manner.
This allows the definition of the previous ``open the fridge'' example with two constraints, hold the handle and move the door according to its articulation model.
In~\cite{bouyarmane2018quadratic}, a similar idea is explored, although the approach is based entirely on URDF~\cite{urdf}, which is limited in its descriptive capabilities.
The approach proposed in this paper is instead similar to~\cite{kineverse}, by which even complex dependent kinematics can be described.
Both are built on top of \casadi{}~\cite{casadi}, a library for computer algebra and automatic differentiation designed specifically for optimal control.

The proposed framework is called \textbf{Giskard} and uniquely combines the following components to close the motion execution gap.
\begin{itemize}
\item \textbf{Motion Statecharts} are provided as executable symbolic motion representations, by which online monitoring and automatic semantic annotation of trajectories are enabled.
\item A \textbf{differentiable kinematic world model} is provided as a structured semantic digital twin, by which world-centric task functions are supported.
\item \textbf{\ac{lmpc}-based task-function control} combines active constraints dictated by a \ac{msc} with explicit jerk bounds to prevent discontinuities during task transition.
\end{itemize}
Together these components not only allow motion descriptions via semantic constraints, but also offer real-time semantic feedback during the motion.
This lets the high-level planner implement a closed semantic feedback loop with the motion executive.
It can understand why motions succeed or fail, and adjust plans, instead of treating execution as a black box.

Transferability was validated on eight diverse platforms, as discussed in Chapter \ref{chapt:trans}, ranging from mobile manipulators such as the PR2 to static dual-arm setups.
Minimal robot-specific tuning was required, by which out-of-distribution generalization was demonstrated.

\section{Differentiable Kinematic World Model}
\label{sec:world-model}
\newcommand{\fx}{{_{b}x}}
\newcommand{\fy}{{_{b}y}}

\newcommand{\ox}{{_{o}x}}
\newcommand{\oy}{{_{o}y}}
\begin{figure}[t]
    \centering
    \begin{subfigure}[t]{0.46\columnwidth}
      \centering
      \includegraphics[width=\textwidth]{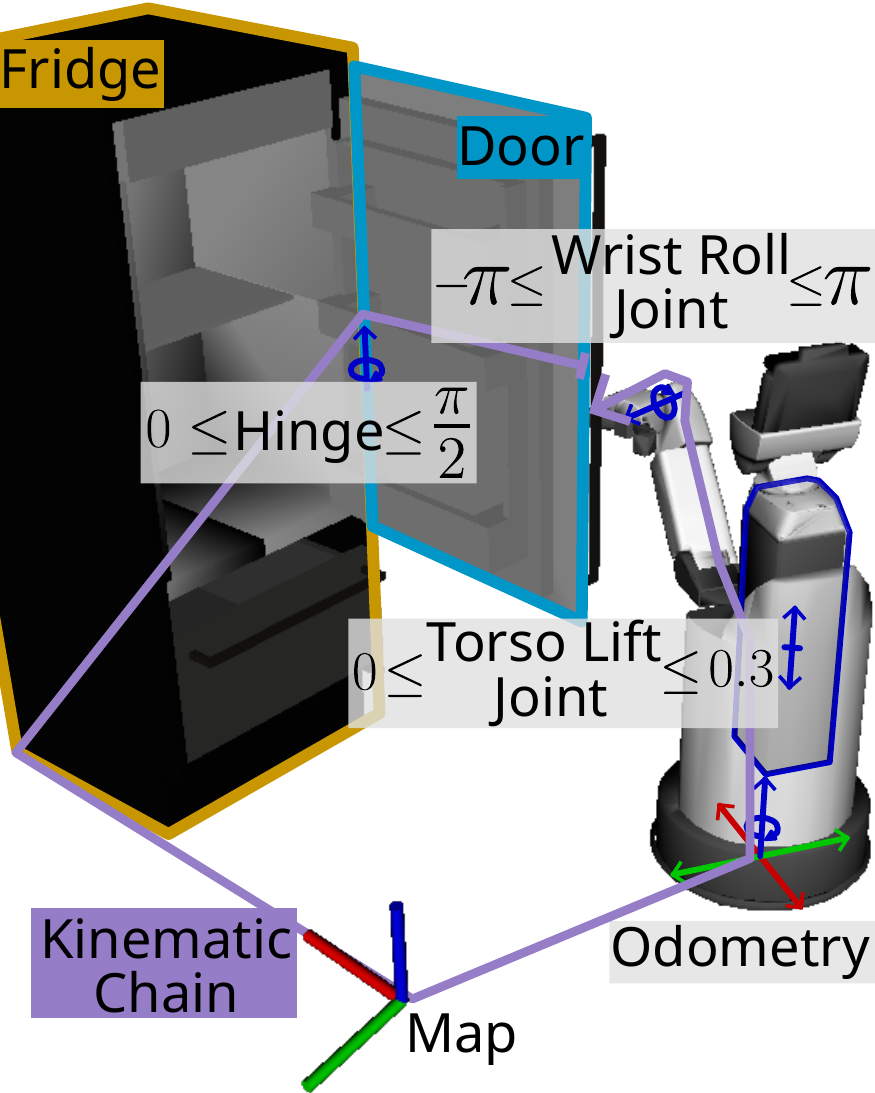}
    \end{subfigure}
    \hfill
    \begin{subfigure}[t]{0.53\columnwidth}
      \centering
      \resizebox{\textwidth}{!}{
        \begin{tikzpicture}[world={5em}{2pt}{2.5em}{3.5em},
        ]
          \node[group={Fridge}{21.5em}, text width=5.25em, xshift=-2.9em, yshift=-2.2em] {};
          \node[group={Robot}{21.5em}, text width=8em, xshift=4.2em, yshift=-2.2em] {};
          \node[link={Map}, name=root] {} [grow'=up]
          child[] { node[joint={Fridge Mount}{0}{0}, xshift=-0.15em] {}
            child[] { node[link={Fridge}] {}
              child[] { node[joint={Hinge}{1}{0}] {}
                child[] { node[link={Door}] {}
                  child[] { node[joint={Handle Joint}{0}{0}] {}
                    child[] { node[link={Handle}] {} }
                  }
                }
              }
            }
          }
          child[] { node[joint={Odometry}{3}{2}, xshift=2.8em] {}
            child[] { node[link={Base Link}] {}
              child[] { node[joint={Torso Lift Joint}{1}{0} xshift=0.4em] {}
                child[] { node[filler] {}
                  child[] { node[joint={Wrist Roll Joint}{1}{0}] {}
                    child[] { node[link={Tool Frame}, yshift=-0.25em] {} }
                  }
                }
              }
              child[] { node[filler, xshift=-1.4em] {}
                child[] { node[link={Head}, text width=2em, yshift=0em] {} }
              }
            }
          }
          ;
        \end{tikzpicture}
      }
    \end{subfigure}
    \caption{An example of a world containing a fridge and a Toyota HSR (left), with its kinematic model (right).}
    \label{fig:world:tree}
\end{figure}
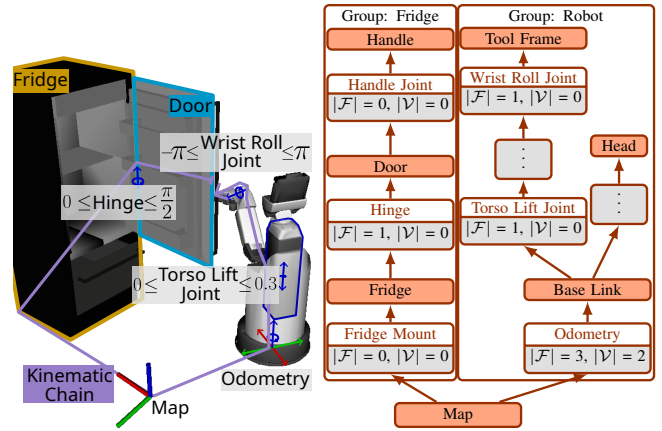

Giskard requires a kinematic model to describe motions with task functions.
An example world model is shown in Fig.~\ref{fig:world:tree}.
The model is defined as follows.
\begin{itemize}
	\item \( \links{} \): A set of links.
	\item \( \joints{} \): A set of joints.
	\item \( \symbols{} = \freeVariables{} \cup \virtualVariables{} \):
    \begin{itemize}
         \item  \( \freeVariables{} \): Symbols referring to \acp{dof}.
     This set includes robot's \acp{dof}, e.g., ``wrist roll joint'', as well as environmental \acp{dof}, e.g., ``hinge''.
      For each such symbol \(\q[]{}\), its time derivatives up to the third order, namely \(\qd[]{}, \qdd[]{}\), and \(\qddd[]{}\) are also included in \(\freeVariables{}\).
    	\item \( \virtualVariables{} \): Symbols representing non-actuated variables, treated as constant by the controller, e.g., robot localization.
    \end{itemize}
	\item \( \worldConstraints{} \): A set of constraints limiting the range of symbols in \(\freeVariables{}\).
\end{itemize}
Links and joints form a tree structure that includes both the robot and the articulated environment.
Joint transformations are defined as functions over \(\symbols{}\).
\begin{equation}
{}^{\texttt{parent}}T_{\texttt{child}} = f(\symbols{}) \in \mathrm{SE}(3)
\end{equation}
They are implemented as \symbolicexpression{} using \casadi{}~\cite{casadi} for automatic differentiation.
A distinction is made between joints and \acp{dof}, since a many-to-many relationship exists between them.
For example, the ``Odometry'' joint contains non-actuated variables to encode positional measurements relative to ``map'' and three \acp{dof} to encode the robot's ability to translate and rotate relative to ``Base Link''.
\acp{dof} of the articulated environment are treated as controllable, for example in the case of a ``Hinge'', since the robot can exert indirect control over them while in contact.
This approach simplifies the definition of environment manipulation tasks, as discussed in the next subsection.

\subsection{Differentiable Task Functions}
\label{subsec:differentiable-task-functions}
\begin{figure}
\includegraphics[width=\columnwidth]{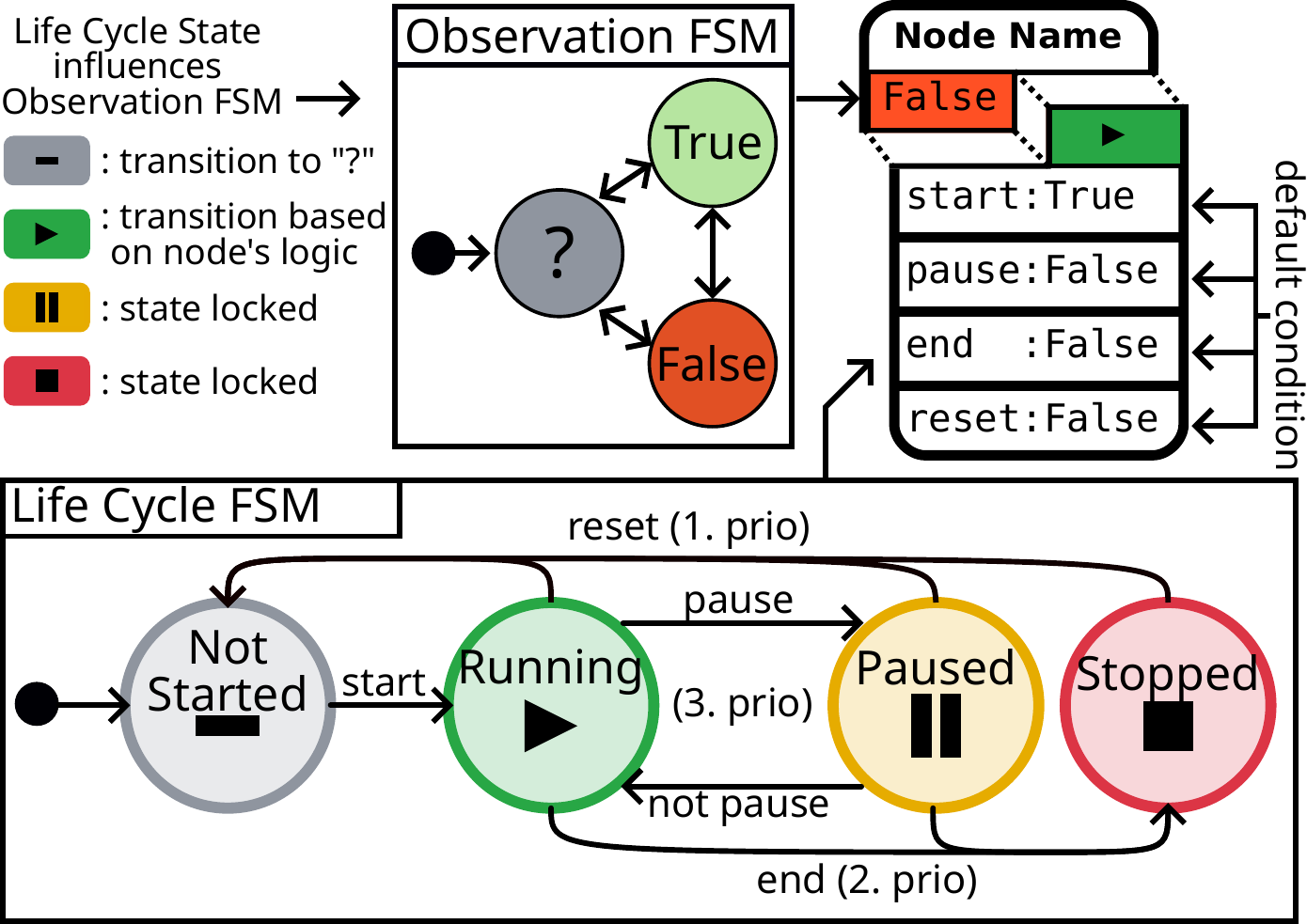}
  \caption{Leaf \ac{msc} node with life cycle \ac{fsm} (bottom) and observation \ac{fsm} (top left): \texttt{True}/\texttt{False}/\texttt{"?"} tracks task achievement for transitions \emph{and} trajectory annotation.}
\label{fig:leaf_node}
\end{figure}
The main purpose of the world model is to provide a means of computing \ac{fk} for chains between any link pairs.
\acp{fk} are computed by multiplying the transformation matrices of all joints along the chain.
Fig.~\ref{fig:world:tree} shows a handle-to-gripper chain spanning both the robot and the environment.
All \ac{fk}s are automatically differentiable thanks to \casadi{} and can be used in the definition of task functions.

The high-level idea of the task function approach is to describe a motion as a set of task functions.
Task functions are constraints defined on differentiable \taskspace{}s.
A \textbf{\taskspace{}} ($\taskF[]{}$) maps the world state onto an arbitrary space:
\[
\taskF[]{}: \mathbb{R}^{|\symbols{}|} \rightarrow \mathbb{R}^{m \times n},
\]
where $m$ and $n$ are arbitrary integers.
The robot must be capable of influencing the \taskspace{}.
Technically, this means that at least one partial derivative with respect to the robot's \acp{dof} is non-zero.
\acp{fk} that include parts of the robot satisfy this definition.
In Fig.~\ref{fig:world:tree}, this enables the description of a door-opening task by combining a joint-space task for the ``Hinge'' with a Cartesian-space task for the gripper to hold the handle.
Furthermore, these expressions can be extended with arithmetic and trigonometric operations provided by \casadi{} to produce arbitrary geometric constructs.

Motions are described by constraining \taskspace{}s using task functions.
A \textbf{Task Function} is an (in-)equality constraint on a scalar-valued \taskspace{}:
\(\taskFEqP = \bE\), or \(\lbASlack \leq \taskFNeqP \leq \ubASlack\).
For example, a 6D Cartesian-space motion goal is defined using six equality constraints, three for position and three for rotation, with \(\taskFEqP\) derived from the \ac{fk} and \(\bE\) being the position/rotation error.
Grasping a handle can be implemented with a \taskspace{} describing the distance between the robot's gripper and a line representing the handle.
A task function can constrain that distance to be equal to zero.
The resulting motion moves the gripper to its closest point on the handle.

\section{Motion Statecharts}
\label{sec:motion-statecharts}
Consider a high-level planner that provides a symbolic action description such as ``open the fridge''.
It may produce a sequence such as ``grasp the handle'' $\rightarrow$ ``move the door open'' $\rightarrow$ ``release the handle'', or more granular steps.
These actions must be grounded into continuous control commands.
Furthermore, to execute this sequence as a single motion, a single set of task functions is insufficient because the constraints for opening the door must not be active during the handle-grasping phase.

Previous work proposes the use of hierarchical \acp{fsm}~\cite{pane2021autonomous,bolotnikova2021task} or \acp{bt}~\cite{dominguez2022stack,diffbrum,rovida2018motion} to group task functions into different phases.
The structure of \acp{bt} makes them difficult to integrate directly into the control loop, as all three cited works run them at lower frequency (2--10 Hz) in parallel with high-frequency controllers.
\acp{fsm} do not share this limitation and support real-time execution, but suffer from the state explosion problem.
As system complexity increases, the number of states grows combinatorially, though hierarchical definitions mitigate this to some extent.

The proposed \acp{msc} build on statecharts to retain \ac{fsm} real-time capability while alleviating state explosion by organizing motion components into a network of dual-\acp{fsm}.
This yields a reusable representation that bridges the motion execution gap by composing motion tasks and monitors into reactive graphs portable across robots and environments.

\subsection{Structure and Elements}

\acp{msc} are defined as a variation of statecharts with explicit motion-related constructs tailored to robot control.
They are represented as graphs whose nodes can contain nested statecharts, by which reusability and structural decomposition are promoted.
All nodes may, but do not have to, be associated with motion task functions (see Sec.~\ref{subsec:differentiable-task-functions}), which remain active for the duration of the node’s activation.
Nodes that are associated with task functions are referred to as \textbf{Motion Tasks}, those without are called \textbf{Monitors}.
An example node is depicted in Fig.~\ref{fig:leaf_node}.
Each node contains two internal \acp{fsm}.
\begin{itemize}
    \item An observation \ac{fsm} (Fig.~\ref{fig:leaf_node}, top-center) is used to represent whether an observed condition is currently \texttt{True}, \texttt{False}, or \texttt{"?"}.
    For Motion Tasks, this state represents the constraint satisfaction status.
    An explicit \texttt{"?"} state is required for situations in which the state is unknown, for example when the node has never been active.
    \item A life cycle \ac{fsm} (Fig.~\ref{fig:leaf_node}, bottom) is used to determine whether the node is active.
\end{itemize}
Conditions are depicted in the lower half of a node (see Fig.~\ref{fig:leaf_node}, top right).
They govern transitions between life cycle states, for example transitioning a node from \taskInactive{} to \taskActive{} when the start condition is satisfied.
These transitions are expressed as ternary logical expressions that combine the three-valued observation states of any node on the same hierarchy level.
In addition, all nodes may implement callbacks that are triggered during every control cycle in which the node is \taskActive{}, or when a transition between life cycle states occurs.
\begin{figure}
    \includegraphics[width=\columnwidth]{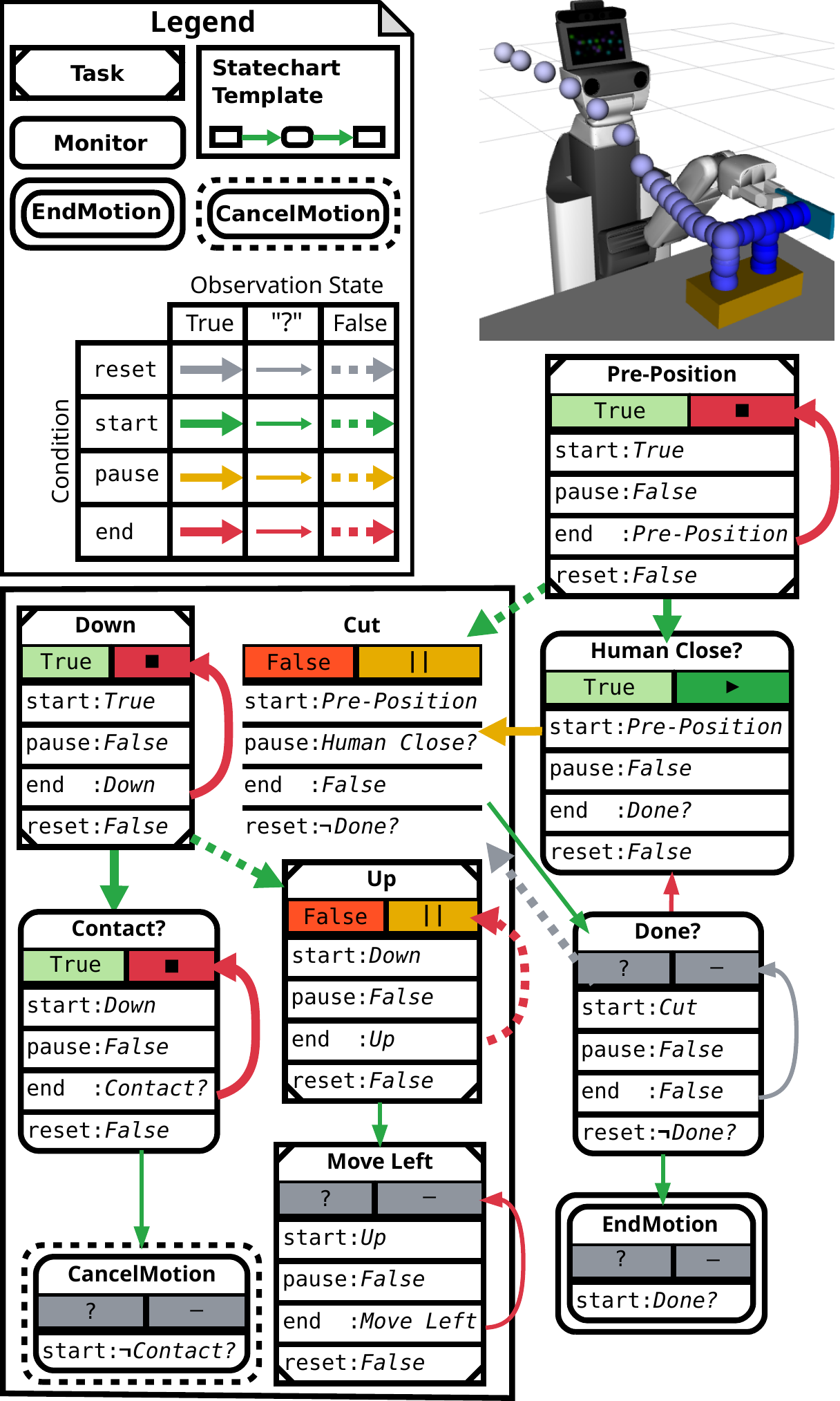}
    \caption{An example \ac{msc} for executing a cutting motion, showcasing all its features.
    It shows a snapshot when a ``Human Close?'' was observed, pausing the ``Cut'' \ac{msc} template.}
    \label{fig:hsr_cut}
\end{figure}
A nested statechart is referred to as a \textbf{\ac{msc} Template}.
\ac{msc} Templates are themselves nodes with the same internal \acp{fsm} structure and contain isolated statecharts.
The life cycle states of nested \ac{msc} nodes are linked to the life cycle state of the parent.
If the parent node is in \taskActive{} state, the child nodes can change state freely.
In all other cases the child nodes mirror the parent's state, if the diagram in Fig.~\ref{fig:leaf_node} allows the transition.
This means a parent cannot force a child transition from \taskInactive{} to \taskOnHold{}, but it can make a child transition from \taskActive{} to \taskDone{}.
In addition to enabling reuse of frequently occurring \ac{msc} structures, templates are used to impose structure on motions.
Common examples include the following.
\texttt{Sequential}: Child nodes are activated upon the predecessor reaching the \texttt{True} state, and the template succeeds when the final child succeeds.
\texttt{Parallel}: All child nodes are activated simultaneously, and the template succeeds when $n$ children reach the \texttt{True} state.

The terminal states of \acp{msc} are referred to as \textbf{EndMotion} and \textbf{CancelMotion}.
Both states trigger termination of the control loop and stop motion execution.
The distinction between the two states is purely semantic feedback.
\begin{figure}
    \includegraphics[width=\columnwidth]{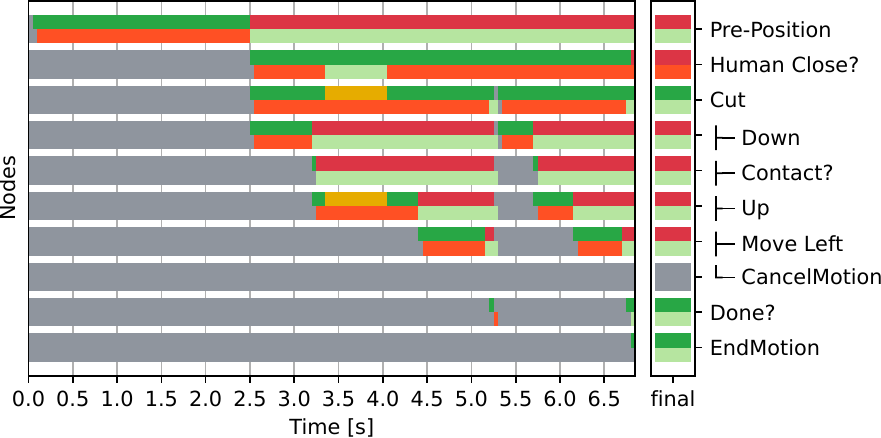}
    \caption{The semantically annotated trajectory corresponding to Fig.~\ref{fig:hsr_cut} as a gantt chart.
    The top bar shows life cycle and bottom bar shows observation states.
    The colors correspond to Fig~\ref{fig:leaf_node}.}
    \label{fig:cut_gantt}
\end{figure}

\subsection{Motion Statecharts in the Control Loop}
\label{subsec:example:-cutting}

At runtime, the state of the \ac{msc} is evaluated at every control cycle in the following order:
\begin{enumerate}
\item The \texttt{on\_tick} callbacks of all \taskActive{} nodes are executed given the current system state, including joint states, transforms, and sensor inputs, in order to update their observation states.
\item All life cycle state machines are updated simultaneously based on the current observation states of all nodes.
\item If a final state exhibits a \texttt{True} observation state, execution terminates and zero commands are sent to the robot.
\item Otherwise, the task functions associated with all \taskActive{} nodes are collected and applied as constraints to the \ac{lmpc}.
\item The \ac{lmpc} computes the next jerk commands for all \acp{dof} of the system, which are integrated to velocity commands and sent to the robot.
\end{enumerate}

\subsection{Example: MSC of a Cutting Motion}

Figure~\ref{fig:hsr_cut} illustrates an \ac{msc} executing a cutting motion on the HSR platform.
The motion starts with the ``Pre-Position'' task, the only node whose start condition is \texttt{True}.
The ``Cut'' \ac{msc} template then coordinates downward, upward, and leftward knife motions through three self-terminating tasks.
The embedded monitor ``Contact?'' aborts the sequence if it fails to activate following ``Down'' completion.
Upon termination of ``Move Left'', the template advances to a \texttt{True} observation state.
The ``Done?'' monitor subsequently determines whether to reset the ``Cut'' template or end execution.

In parallel, the ``Human Close?'' monitor remains active, pausing the ``Cut'' template on human proximity detection, as depicted in the figure snapshot.
At this time, ``Down'' and ``Contact?'' have already reached \taskDone{}.
These nodes remain in \taskDone{} because no transition to \taskOnHold{} exists (cf. Fig.~\ref{fig:leaf_node}).
``Up'' transitions from \taskActive{} to \taskOnHold{}, because such a transition is possible.
Unstarted child nodes remain in \taskInactive{}.

Node observation states serve dual purposes:
triggering life cycle transitions and enabling \textit{real-time motion annotation}.
Figure~\ref{fig:cut_gantt} presents a Gantt chart of the cutting motion execution, with each node depicted by dual bars representing life cycle and observation states, respectively.
Human proximity is evident between $t\approx3.4$ and $t\approx4.1$ via the monitor's \texttt{True} observation state.

\section{lMPC-Based Implementation of the Task Function Approach}
\label{sec:qp}
\acp{msc} allow task functions to be exchanged at every control cycle.
In standard \ac{qp}-controllers, this can introduce discontinuities.
Alternatives defer switching until motion phase end~\cite{dominguez2022stack} or use weight scaling~\cite{pane2021autonomous}.
The former limits motions to separable phases with pauses, the latter is unable to deactivate constraints immediately and adds tuning parameters for the weight-scaling strategies.

To address this, a \ac{lmpc}-based task function implementation was developed that prevents discontinuities in the joint velocity commands using jerk limits.
It predicts system evolution over a short horizon to enforce jerk limits, with task functions as constraints.

\subsection{Formulation as a Quadratic Program}
\begin{figure*}[t]
    \centering
    \includegraphics[width=0.925\textwidth]{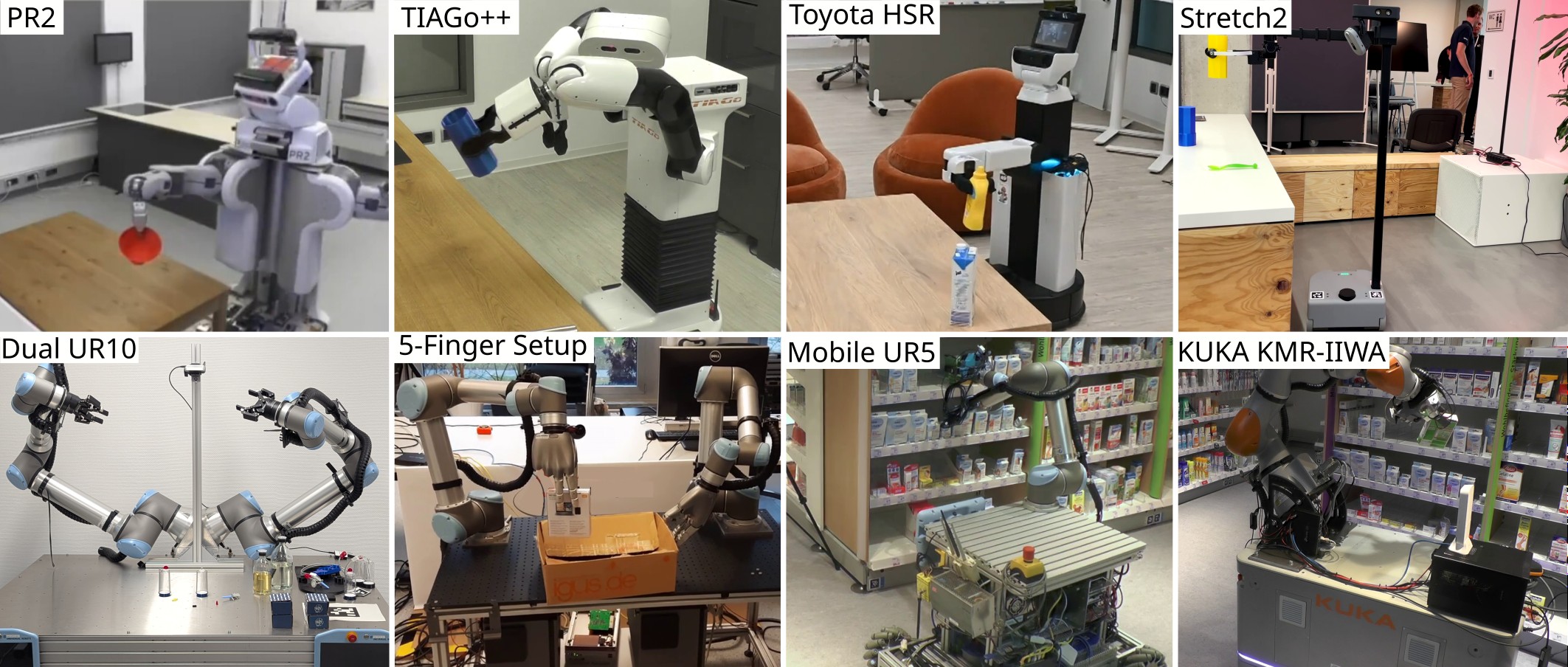}
    \caption{
    The eight real robots the proposed framework has been deployed on.
    }
    \label{fig:6-robots}
\end{figure*}

The \ac{lmpc} is formulated as a \ac{qp}:
\begin{small}
\begin{align}\label{eq:main}
    \min\limits_{\qd[v], \qddd[v], \slackEqP[v], \slackNeqP[v]} \quad &
    \sum_{\timecounter=0}^{\PredictionHorizon-3}(
    \|\qd[v]_{\timecounter}\|_{\wqd[v][, \timecounter]{}}^2
    )
    + \|\slackEqP[v]\|_{\wEqP[v]}^2 
    + \|\slackNeqP[v]\|_{\wNeqP[v]}^2 \quad & \quad
\end{align}
\begin{align}
    \text{s.t.} \quad
    &\left.
        \begin{matrix}
            \lbAVel[v][,\timecounter]{} \leq \qd[v]{}_{\timecounter} \leq \ubAVel[v][,\timecounter]{} \\
            \qddd[v]{}_{\min} \leq \qddd[v]{}_{\timecounter} \leq \qddd[v]{}_{\max} \\
        \end{matrix}
    \right.  \quad
    \begin{array}{l}
        \forall \timecounter = 0, \ldots, \PredictionHorizon - 3 \\
        \forall \timecounter = 0, \ldots, \PredictionHorizon - 1
    \end{array}\text{\ (bounds)}\label{eq:box-constriants}\\
    &\left.
        \begin{matrix}
            {\qd[v]{}}_{0} = {\qd[v]{}}_{\mathrm{curr}} + {\qdd[v]{}}_{\mathrm{curr}} \dt + \qddd[v]{}_{0} \dt^2 \\
            {\qd[v]{}}_{1} = 2{\qd[v]{}}_{0} - {\qd[v]{}}_{\mathrm{curr}} + \qddd[v]{}_{1} \dt^2 \\
            {\qd[v]{}}_{\timecounter} = 2{\qd[v]{}}_{\timecounter - 1} - {\qd[v]{}}_{\timecounter - 2} + \qddd[v]{}_{\timecounter} \dt^2 \\
            0 = 2{\qd[v]{}}_{\PredictionHorizon - 3} - {\qd[v]{}}_{\PredictionHorizon - 4} + \qddd[v]{}_{\PredictionHorizon - 2} \dt^2 \\
            0 = - {\qd[v]{}}_{\PredictionHorizon - 3} + \qddd[v]{}_{\PredictionHorizon - 1} \dt^2 \\
        \end{matrix}
    \right\}
    \begin{array}{l}
       (\text{system model})\\
        \forall k = 1, \dots, \PredictionHorizon - 3\\
       \text{  }\\
    \end{array}\label{eq:system-model}\\
&
        \bE[v] = \dt \sum\limits_{\timecounter=0}^{\PredictionHorizon - 3} \Jacobian[][][\taskFEqP[v]{}][]{} \qd[v]{}_{\timecounter} + \dt \slackEqP[v] \quad\quad\quad \text{(eq. task func.)} \label{eq:eq-task-functions}\\
        & \lbASlack[v] \leq \dt \sum\limits_{\timecounter=0}^{\PredictionHorizon - 3} \Jacobian[][][\taskFNeqP[v]{}][]{} \qd[v]{}_{\timecounter} + \dt \slackNeqP[v] \leq \ubASlack[v] \quad \text{(neq. task func.)}
    \label{eq:neq-task-functions}
\end{align}
\end{small}

The first term of the objective function minimizes the joint velocities $\qd[v]{}_{\timecounter}$, weighted by $\wqdnoT[v]{}$, over a prediction horizon $\PredictionHorizon$.
They are represented as vectors containing one variable per \ac{dof} of the system.
$\PredictionHorizon$ is chosen to be sufficient to enforce jerk limits, with $\PredictionHorizon \dt$ typically below $\leq 0.25$s.
The weights start at \(0.001\) and increase linearly to \(0.01\) at the end of the horizon, thereby encouraging the use of immediate velocities.

The remaining objective function minimizes vectors of slack variables \(\slackEqP[v]\) and \(\slackNeqP[v]\), associated with equality and inequality constraints, respectively.
They prevent infeasibility in the presence of conflicting constraints through relaxation.
By default, their weight is set to \(1\), normalized by the square of the expected maximum velocity of the corresponding task.
As a consequence, using joint velocity to fulfill constraints is approximately two orders of magnitude cheaper than violating them using slack variables.

The jerk decision variables $\qddd[v]$ of the objective function influence it only indirectly.
Instead, the \textbf{system model} links velocity and jerk through constraints derived from a semi-implicit Euler integration scheme,
\begin{align*}
    {\qd[v]{}}_{\timecounter{}} &= {\qd[v]{}}_{\timecounter{} - 1} + {\qdd[v]{}}_{\timecounter{}} \dt, \\
    {\qdd[v]{}}_{\timecounter{}} &= {\qdd[v]{}}_{\timecounter{} - 1} + {\qddd[v]{}}_{\timecounter{}} \dt.
\end{align*}
Consequently, the weights applied to velocity variables implicitly influence the jerk variables.
The system model in Eq.~\ref{eq:system-model} is obtained by eliminating acceleration variables, enforcing a terminal velocity of zero for stability, and linking the current states to the initial system state.
Explicit acceleration variables are omitted because their bounds had negligible impact in practice.
This reduces the number of decision variables by roughly $\frac{1}{3}$.
Introducing explicit jerk penalties to the objective function can destabilize the system by making continued motion cheaper than deceleration.

The \textbf{bounds} defined in Eq.~\ref{eq:box-constriants} impose direct limits on velocities and jerks.
Given the prediction horizon and desired velocity limits, a theoretical maximum jerk exists that allows the system to decelerate from the velocity limit within the horizon.
This value is computed once by solving a modified version of Eq.~\ref{eq:main} without jerk bounds.
Empirically, no benefit was observed from further manual tuning of jerk limits.
As a result, $\PredictionHorizon$ and $\dt$ are the only parameters requiring robot-specific tuning.
Joint position limits are encoded into the velocity bounds by progressively restricting allowed values as joints approach their limits.

Finally, \textbf{task functions} are incorporated into the final two constraint sets in Eq.~\ref{eq:eq-task-functions} and \ref{eq:neq-task-functions} by constraining the Jacobian integral over the horizon and relaxing it with slack variables.
In Eq.~\ref{eq:eq-task-functions} $\bE[v]{}$ corresponds to the \taskspace{} error.
In Eq.~\ref{eq:neq-task-functions} $\lbASlack[v]{}$ and $\ubASlack[v]{}$ define lower and upper bounds on the \taskspace{} error, respectively.
To prevent excessively large error terms, all three bounds are clamped to values achievable within the horizon.
The slack variables $\slackEq[v]$ and $\slackNeq[v]$ relax these constraints as described previously.
They represent the residual \taskspace{} velocity integral over the horizon and are scaled by $\dt$ such that their magnitudes are comparable to the integral term.
This formulation requires only a single constraint and a single slack variable per task function in the \ac{qp}.
The majority of motions encountered in mobile manipulation can be expressed using such positional task functions, which supports scalability of the overall system.
Velocity task constraints can also be formulated, but they require one constraint per discretization step of the prediction horizon.
Further details on this formulation are provided in~\cite{stelter25giskard}.

For this application, the fastest available \ac{qp} solver is qpSWIFT~\cite{qpswift}, which enables execution of the complete control loop at 100 hz on a high-performance CPU.

\section{Evaluation}
\label{sec:evaluation}
The presented motion executive aims to close the motion execution gap, by offering a method to describe motions semantically with \ac{msc} and translating it into a kinematic \ac{lmpc} for a given robot.
The framework was deployed on eight robot platforms and several different environments, shown in Fig.~\ref{fig:6-robots}.

\subsection{Closing the Motion Execution Gap with MSC}
To test if the presented framework is able to close the motion execution gap, it was integrated as the motion executive into the CRAM cognitive architecture~\cite{beetz2023cram}.
This architecture uses a knowledge base to specialize its plans and parameterize motion designators, which serve as the symbolic counterpart to \acp{msc} in our framework.
To integrate our approach, the high-level planner was modified to resolve plans to \acp{msc} instead.

Returning to the cutting example (Fig.~\ref{fig:cut_real}), a task such as ``cut the zucchini'' does not yet specify how the cutting should be carried out.
Queries to a knowledge base provide answers, for instance that the zucchini should be sliced instead of cubed, or what the preceding and following motions are.
The answers allow the decomposition of the plan into sub-actions and finally \acp{msc}, linking them to their semantic meaning.
This setup allows the high-level planner to reason about the current system state during execution based on the observation and life cycle states of nodes.
After the execution, it can use the final \ac{msc} state to reason about failure causes.

The high-level planner decides the next step based on this semantic feedback.
It may stop the plan, repeat a failed cut, switch tools, or change settings like the target position or direction of movement.
The cutting process is described by a discrete separation state $S(t)\in\{\texttt{False},\texttt{True}\}$, where $S(t)=\texttt{True}$ means that material separation has taken place at time $t$.
A \ac{msc} monitor tracks this semantic state using geometric and contact observations.
Another monitor tracks whether the contact force at the tool frame is increasing: $C(t)=\dot{\mathbf{f}}(t) > 0$.
This feedback is interpreted on the high-level by evaluated logical conditions.
For example ``CutInProgress'' is defined as both monitors having a \texttt{True} observation state:
$
C(t) \;\land\; S(t)
\;\Rightarrow\; \text{CutInProgress}.
$
Under this model, cutting inside the high-level planner is no longer described using a specific method or perception routines.
Instead the high-level decisions are made based on semantic feedback.
This clear mapping of physical signals to semantic states lets us reason, for example, that ``the knife did not touch the object'' if no contact is detected.

\subsection{Transferability}
\label{chapt:trans}
\begin{figure*}[t]
    \centering
    \includegraphics[width=\textwidth]{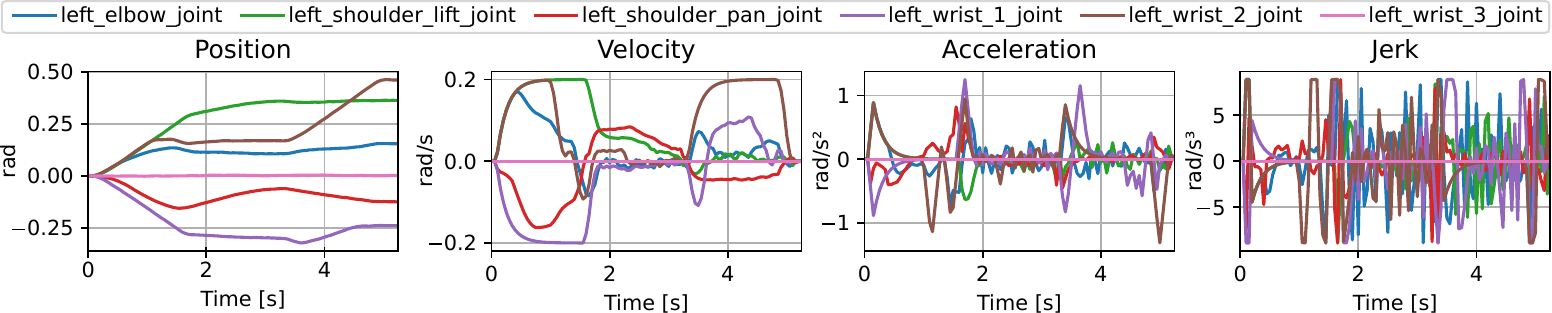}
    \caption{Joint-space trajectory of the insertion motion corresponding to Fig.~\ref{fig:tracy_insert}, recorded during real-robot execution.
    Joint position values are offset to originate at $0$ for readability.}
    \label{fig:insert-traj}
\end{figure*}
\begin{table}[ht]
\centering
\small
\begin{tabular}{lll}
\toprule
\ac{msc} Node & Real Robots & Kin. Sim. Tests \\
\midrule
Cartesian pose & 8/8 & 8/8 \\
Joint goal & 8/8 & 8/8 \\
Feature Functions & 8/8 & 8/8 \\
Door/Drawer opening & PR2, TIAGo, HSR & 6/8 (mobile only) \\
Cutting & PR2 & 8/8 \\
Diff-drive nav & TIAGo, Stretch2 & TIAGo, Stretch2 \\
Peg-In-Hole & Dual UR10 & 8/8 \\
5-finger grasp & 5-Finger Setup & 5-Finger Setup\\
\bottomrule
\end{tabular}
\caption{List of all implemented \ac{msc} nodes and the real robots they have been used on.
Furthermore, all combinations that should be possible based on a robot's kinematics were verified in simulation.}
\label{tab:transfer}
\end{table}
To showcase the transferability of the framework, Table~\ref{tab:transfer} list all implemented \ac{msc} nodes and the real robots they have been deployed on.
Since our approach uses a kinematic world model, all implementations are, in principle, robot-agnostic and can be reused on any platform that offers the required capabilities (e.g., a differential drive for the ``Diff-drive navigation'' node).
To verify this, the world model presented in Section~\ref{sec:world-model} was used to kinematically simulate sensible combinations.

``Feature Functions'' are defined as algebraic expressions that combine geometric primitives, such as points, lines, and planes, between tool and object frames to form differentiable scalar constraints~\cite{featurefunctions}.
They enable the specification of complex motions through composition within \ac{msc}.
``Diff-drive navigation'' addresses the limitation of platforms such as TIAGo++ and Stretch2, which are unable to move sideways.
Corresponding templates sequence orientation, forward driving, and reorientation actions to replace Cartesian goals for driving.

The tuning effort required to achieve this transferability was minimal.
\acp{msc} themselves do not require any robot-specific tuning.
The \ac{lmpc} formulation depends on only two parameters, $\dt$ and $\PredictionHorizon$ (see Sec.~\ref{sec:qp}).
The former, $\dt$, is matched to the joint feedback rate and typically lies between $0.01$s and $0.02$s.
Increasing $\PredictionHorizon$ improves motion smoothness at the cost of increased computational effort.
This parameter was selected as the minimum value required to accommodate robot-specific noise levels, typically between 15 and 30.
For kinematic simulation, identical parameters were used across all setups, namely $\dt = 0.02$s and $\PredictionHorizon = 7$, in order to minimize computation time.
The resulting level of transferability exceeds what can be achieved by single generative models without platform-specific fine-tuning.


\subsection{Smooth Motion Task Transition}

\begin{figure}[t]
    \centering
    \includegraphics[width=0.49\columnwidth]{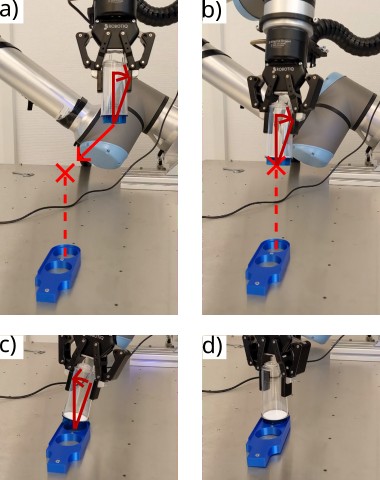}
    \includegraphics[width=0.49\columnwidth]{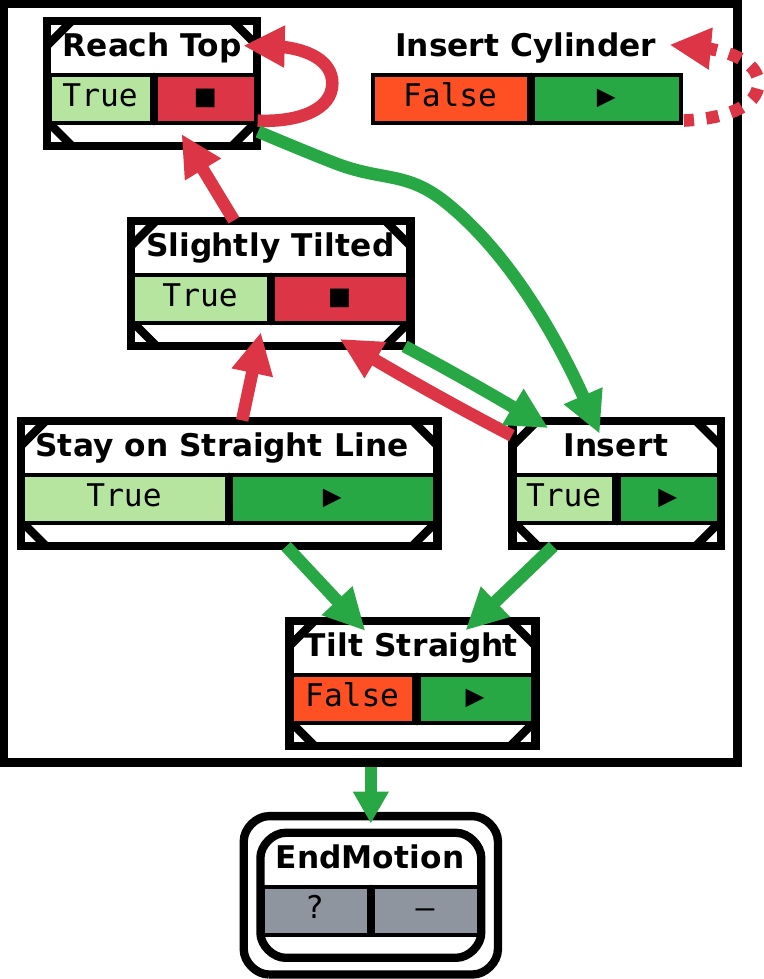}
    \caption{
    The dual UR10 setup executing an insertion motion and the corresponding \ac{msc} at the time of image~c).
    Life cycle transitions of nodes are omitted to conserve space.
    }
    \label{fig:tracy_insert}
\end{figure}

The dual UR10 setup was deployed in a sterility testing use case.
One of the required motions was the insertion of a cylindrical object into a slot, corresponding to a classical peg-in-hole task.
Using feature functions, an \ac{msc} template was defined in which the motion is decomposed into multiple phases, as illustrated in Fig.~\ref{fig:tracy_insert}.

\begin{itemize}
\item \textbf{Always}: ``Stay on Line'', by which the distance between the cylinder bottom and the hole axis is minimized.
\item \textbf{Approach}: ``Reach Top'', minimizing the distance between the cylinder bottom and a point above the hole, combined with ``Slightly Tilted'', enforcing that the angle between the cylinder and the horizontal world axis remains above a specified threshold.
\item \textbf{Insert}: The tilt constraint is replaced by ``Insert'', formulated as a point-to-point goal.
\item \textbf{Finish}: ``Tilt Straight'', aligning the cylinder axis with the hole axis.
\end{itemize}

The resulting execution demonstrates that task functions can be exchanged seamlessly during runtime.
The measured trajectory shown in Fig.~\ref{fig:insert-traj} exhibits smooth velocity profiles despite frequent task switches and the presence of sensor noise.
This behavior empirically validates the proposed \ac{lmpc}-based formulation for smooth motion task transitions.

\section{Conclusion}
This paper closes the motion execution gap by introducing \ac{msc} for symbolic motion composition, differentiable kinematic world models as semantic digital twins, and jerk-bounded \ac{lmpc} for smooth task-function control.
Validated on eight diverse robots like PR2 or Stretch2, it enables world-centric specifications, online monitoring, and robust transferability with guarantees unattainable by generative AI, which fails to enforce constraints or long-horizon consistency.

\section{Limitations \& Future Work}
The framework currently lacks dedicated force control support.
Basic force-aware behaviors can nonetheless be realized via monitors that react to sensor feedback and trigger \ac{msc} transitions, as demonstrated in the cutting example.
The reason is that the kinematic world model does not yet represent robot dynamics, though \ac{qp}-based controllers that handle such effects are demonstrated in related work~\cite{bolotnikova2021task}.

Finally, the system focuses on motion execution rather than planning, leaving integration with sampling based planners or trajectory optimizers as future work.

\newpage

\section*{Acknowledgments}
This work was partially supported by: (1) the German Research Foundation DFG, as part of Collaborative Research Center (Sonderforschungsbereich) 1320 Project-ID 329551904 "EASE - Everyday Activity Science and Engineering", University of Bremen (\url{http://www.ease-crc.org/});
(2) the German Federal Ministry of Research, Technology and Space (BMFTR) under the Robotics Institute Germany (RIG);
(3) the 'FAME' project, supported by the European Research Council (ERC) via the HORIZON ERC Advanced Grant no. 101098006.
While partly funded by the European Union, views and opinions expressed are however those of the authors only and do not necessarily reflect those of the European Union.
Neither the European Union nor the granting authority can be held responsible for them.

\bibliographystyle{named}
\bibliography{ijcai26}

\end{document}

%% file: cmds/names.tex
\newtcbox{\myredbox}[1][]{
    on line,
    colback=myred1, 
    boxrule=0mm, 
    arc=2mm,       
    boxsep=0mm,    
    left=0.5mm,      
    right=0.5mm,     
    top=0.5mm,     
    bottom=0.5mm,  
    #1
}
\newtcbox{\myyellowbox}[1][]{
    on line,
    colback=myyellow1, 
    boxrule=0mm, 
    arc=2mm,       
    boxsep=0mm,    
    left=0.5mm,      
    right=0.5mm,     
    top=0.5mm,     
    bottom=0.5mm,  
    #1
}
\newcommand{\todo}[1]{\textcolor{red}{TODO: #1\\}}

\newcommand{\symbolicexpression}{symbolic expression}
\newcommand{\casadi}{CasADi}
\newcommand{\taskspace}{task space}

%% file: cmds/tikz.tex
\usetikzlibrary{
  trees,
  positioning, 
  arrows,       
  fit,         
  shapes,       
  decorations.markings,        
  decorations.pathmorphing,
  bending,     
  calc,
}
\usetikzlibrary{arrows.meta,      
                positioning,
                shadows, 
                shapes.multipart}  
\newcommand{\triton}[1]{%
  \tikzpicture[rounded corners = 0pt, fill = none, {}-{}]
    \draw[thick, #1]
      (0, 0.07) -- (0, -0.07)
      (0, 0) -| (-0.1, -0.07)
      (0, 0) -| (0.1, -0.07);
  \endtikzpicture
}
\newcommand\circledarrow[1]{
  \tikzpicture[baseline=-1.4ex, {}-{>[line width=0.5pt, width=0.1cm 1, bend]}]
    \draw[thick, #1]
      (0,0) arc(90:-270:0.7ex);
  \endtikzpicture
}
\newcommand{\puzzleedgeorflip}[4]{
  \tikzpicture[scale = 1, {}-{}]
    \draw[#2, line width=3pt] (0.2,0.0) -- (0.8,0.0);
    \draw[#4, line width=2pt, fill=#2]
          (0.1,#1*0.00) .. controls (0.1,#1*0.00) and (0.3,#1*-0.04) .. 
          (0.3,#1*0.04) .. controls (0.3,#1*0.11) and (0.2,#1*0.20) .. 
          (0.5,#1*0.20) .. controls (0.8,#1*0.20) and (0.7,#1*0.11) .. 
          (0.7,#1*0.04) .. controls (0.7,#1*-0.04) and (1.0,#1*0.00) .. 
          (1.0,#1*0.00);
    \draw[#3, line width=3pt] (1.2,0.0) -- (1.8,0.0);
    \draw[#4, line width=2pt, fill=#3]
          (1.0,#1*0.00) .. controls (1.0,-#1*0.00) and (1.3,-#1*-0.04) .. 
          (1.3,-#1*0.04) .. controls (1.3,-#1*0.11) and (1.2,-#1*0.20) .. 
          (1.5,-#1*0.20) .. controls (1.8,-#1*0.20) and (1.7,-#1*0.11) .. 
          (1.7,-#1*0.04) .. controls (1.7,-#1*-0.04) and (2.0,-#1*0.00) .. 
          (2.0,-#1*0.00);
    \draw[#2, line width=3pt] (2.2,0.0) -- (2.8,0.0);
    \draw[#4, line width=2pt, fill=#2]
          (2.0,-#1*0.00) .. controls (2.0,#1*0.00) and (2.3,#1*-0.04) .. 
          (2.3,#1*0.04) .. controls (2.3,#1*0.11) and (2.2,#1*0.20) .. 
          (2.5,#1*0.20) .. controls (2.8,#1*0.20) and (2.7,#1*0.11) .. 
          (2.7,#1*0.04) .. controls (2.7,#1*-0.04) and (2.9,#1*0.00) .. 
          (2.9,#1*0.00);
  \endtikzpicture
}
\newcommand{\puzzleedgesex}{\puzzleedgeorflip{-1}{white}{background}}
\tikzset{%
  bt/minimumwidth/.initial = 5em,
  bt/minimumheight/.initial = 2pt,
  bt/.default = {5em}{2pt},
  bt/.style n args={2}{%
    font = \scriptsize,
    align = center, text = black,
    draw = foregroundros, fill = background, line width=1pt,
    every path/.style = {color = foregroundros, text = black, line width=1pt},
    {}-{latex}, 
    bt/minimumwidth = #1,
    bt/minimumheight = #2,
    level distance=1.5em, 
    sibling distance=#1+0.5em, 
    growth parent anchor = south,
    edge from parent path={(\tikzparentnode) -- (\tikzchildnode.north)},
    block/.style = {%
      draw = foregroundros, fill = background,
      line width = 1pt, rounded corners = 2pt,
      minimum height = \pgfkeysvalueof{/tikz/bt/minimumheight},
      inner ysep = 2pt, inner xsep = 2pt, inner sep = 2pt,
      text width = \pgfkeysvalueof{/tikz/bt/minimumwidth}, align = center,
      anchor = mid, 
    },
    contents/.style = {local bounding box = everything},
    toplevel/.style = {block,
      fill = none, draw, thin,
      yshift = 6pt, inner ysep = 16pt, inner xsep = 10pt,
      label = {[label distance=-15pt]90:##1},
      fit = (everything),
    },
    contentstwo/.style = {local bounding box = ##1},
    topleveltwo/.style n args=2{block,
      fill = none, draw, thin,
      yshift = 6pt, inner ysep = 16pt, inner xsep = 10pt,
      label = {[label distance=-15pt]90:##1},
      fit = (##2),
    },
    begin/.style = {%
      circle, fill = foregroundros},
    end/.style = {begin,
      double, very thick, double distance = 1pt, draw = foregroundros,
    },
    action/.style = {block,
      node contents = {\textcolor{black}{##1}},
      draw = foregroundros, fill = myyellow1,
    },
    condition/.style = {block,
    node contents = {\textcolor{black}{? ##1}},
    draw = foregroundros, fill = myyellow1,
    },
    lcfsmNotStarted/.style = {block,
      baseline=(current bounding box.center),
      path picture = {
        \draw[thick, rounded corners=0pt, draw=black, fill=black] (0.1*\pgfkeysvalueof{/tikz/bt/minimumwidth},0.03) rectangle 
        (0.9*\pgfkeysvalueof{/tikz/bt/minimumwidth},-0.03);
      },
      draw = lcfsmNotStartedColor, fill = lcfsmNotStartedColor,
    },
    lcfsmRunning/.style = {block,
      node contents = {\tiny{\textcolor{black}{$\blacktriangleright$}}},
      draw = lcfsmRunningColor, fill = lcfsmRunningColor,
    },
    lcfsmPaused/.style = {block,
      path picture = {
        \draw[thick, rounded corners=0pt, draw=black, fill=black] 
        (0.1*\pgfkeysvalueof{/tikz/bt/minimumwidth},0.1) rectangle 
        (0.35*\pgfkeysvalueof{/tikz/bt/minimumwidth},-0.1);
        \draw[thick, rounded corners=0pt, draw=black, fill=black] 
        (0.65*\pgfkeysvalueof{/tikz/bt/minimumwidth},0.1) 
        rectangle 
        (0.9*\pgfkeysvalueof{/tikz/bt/minimumwidth},-0.1);
      },
      draw = lcfsmPausedColor, fill = lcfsmPausedColor,
    },
    lcfsmStopped/.style = {block,
      path picture = {
        \draw[thick, rounded corners=0pt, draw=black, fill=black] 
        (0.15*\pgfkeysvalueof{/tikz/bt/minimumwidth},0.06) rectangle 
        (0.85*\pgfkeysvalueof{/tikz/bt/minimumwidth},-0.06);
      },
      draw = lcfsmStoppedColor, fill = lcfsmStoppedColor,
    },
    composite/.style 2 args={block,
      rectangle split, rectangle split parts = 2,
      fill = none, rectangle split part fill={none, background},
      text height = 0.5em,
      text = foregroundros,
      inner ysep = 0.1em,
      node contents = {
        {\boldmath ##1} 
        \nodepart[inner ysep = 0pt,inner xsep = 0pt]{second} \textcolor{black}{##2}\vspace{0.25em}
      },
    },
    async/.style={
      composite={$\pmb{\rightsquigarrow}$}{##1},
    },
    sequence/.style = {composite={$\pmb{\rightarrow}$}{##1},
    },
    parallel/.style =  {
      composite={$\pmb{\rightrightarrows}$}{##1},
      text depth = 0.5pt,
    },
    selector/.style =  {
      composite={\textbf{?}}{##1},
    },
    loop/.style = {
      composite={\circledarrow{darkredaccent}}{##1}, 
    },
    subtree/.style = {composite={\triton{darkredaccent}}{##1},
    },
    highlight/.style = {
      draw = darkgreen, fill = lightgreen,
    }, 
    puzzlemale/.style = {
      label = {[scale=0.5, yshift=1pt, anchor=center]-90:{\puzzleedgeorflip{-1}{background}{white}{foregroundros}}},
    },
    puzzlemalehighlight/.style = {
      highlight,
      label = {[scale=0.5, yshift=1pt, anchor=center]-90:{\puzzleedgeorflip{-1}{lightgreen}{white}{darkgreen}}},
    },
    puzzlefemale/.style = {
      label = {[scale=0.5, yshift=-1pt, anchor=center]90:{\puzzleedgeorflip{-1}{white}{white}{foregroundros}}},
    },
    puzzlesex/.style = {
      label = {[scale=0.5, yshift=-2em, anchor=center]90:{\puzzleedgesex}},
    },  
    running/.style = {
      draw = myblue4, fill = myblue1, text = myblue4,
    },    
    success/.style = {
      draw = mygreen4, fill = mygreen1, text = mygreen4,
    },   
    failure/.style = {
      draw = darkred, fill = lightred, text = darkred,
    },
    cross/.style 2 args={
      overlay,
      inner sep=0pt,
      minimum height=##1,
      minimum width=##2,
      yshift=-0.5*##1,
      append after command={
        \pgfextra{
          \draw[black, very thick, -] (\tikzlastnode.south west) -- (\tikzlastnode.north east);
          \draw[black, very thick, -] (\tikzlastnode.north west) -- (\tikzlastnode.south east);
        }
      }
    }
  }
}

\tikzset{%
  world/minimumwidth/.initial = 5em,
  world/minimumheight/.initial = 2pt,
  world/evenleveldist/.initial = 2em,
  world/oddleveldist/.initial = 2em,
  world/.default = {5em}{2pt}{2em}{2em},
  world/.style n args={4}{%
    font = \scriptsize,
    align = center, text = black,
    draw = foregroundlib, fill = background, line width=1pt,
    every path/.style = {color = foregroundlib, text = black, line width=1pt},
    {}-{latex}, 
    world/minimumwidth = #1,
    world/minimumheight = #2,
    world/evenleveldist = #3,
    world/oddleveldist = #4,
    level 1/.style={level distance=\pgfkeysvalueof{/tikz/world/oddleveldist}},
    level 2/.style={level distance=\pgfkeysvalueof{/tikz/world/evenleveldist}},
    level 3/.style={level distance=\pgfkeysvalueof{/tikz/world/oddleveldist}},
    level 4/.style={level distance=\pgfkeysvalueof{/tikz/world/evenleveldist}},
    level 5/.style={level distance=\pgfkeysvalueof{/tikz/world/oddleveldist}},
    level 6/.style={level distance=\pgfkeysvalueof{/tikz/world/evenleveldist}},
    level 7/.style={level distance=\pgfkeysvalueof{/tikz/world/oddleveldist}},
    level 8/.style={level distance=\pgfkeysvalueof{/tikz/world/evenleveldist}},
    level 9/.style={level distance=\pgfkeysvalueof{/tikz/world/oddleveldist}},
    level 10/.style={level distance=\pgfkeysvalueof{/tikz/world/evenleveldist}},
    level 11/.style={level distance=\pgfkeysvalueof{/tikz/world/oddleveldist}},
    level 12/.style={level distance=\pgfkeysvalueof{/tikz/world/evenleveldist}},
    sibling distance=#1+0.5em, 
    edge from parent path={(\tikzparentnode) -- (\tikzchildnode.south)},
    block/.style = {%
      fill = background,
      line width = 1pt, rounded corners = 2pt,
      minimum height = \pgfkeysvalueof{/tikz/world/minimumheight},
      inner ysep = 2pt, inner xsep = 2pt, inner sep = 2pt,
      text width = \pgfkeysvalueof{/tikz/world/minimumwidth}, align = center,
      anchor = mid, 
    },
    contents/.style = {local bounding box = everything},
    toplevel/.style = {block,
      fill = none, draw, thin,
      yshift = 6pt, inner ysep = 16pt, inner xsep = 10pt,
      label = {[label distance=-15pt]90:##1},
      fit = (everything),
    },
    contentstwo/.style = {local bounding box = ##1},
    topleveltwo/.style n args=2{block,
      fill = none, draw, thin,
      yshift = 6pt, inner ysep = 16pt, inner xsep = 10pt,
      label = {[label distance=-15pt]90:##1},
      fit = (##2),
    },
    begin/.style = {%
      circle, fill = foregroundlib},
    end/.style = {begin,
      double, very thick, double distance = 1pt, draw = foregroundlib,
    },
    link/.style = {block,
      node contents = {##1},
      draw = foregroundlib, fill = myred1,
    },
    condition/.style = {block,
      shape=ellipse, inner sep = 0pt,
      node contents = {##1},
      draw = foregroundlib, fill = myyellow1,
    },
    composite/.style 2 args={block,
      rectangle split, rectangle split parts = 2,
      fill = none, rectangle split part fill={none, background},
      text height = 0.5em,
      text = foregroundlib,
      inner ysep = 0.1em,
      node contents = {
        {\color{foregroundlib} ##1} 
        \nodepart[inner ysep = 0pt,inner xsep = 0pt]{second} \textcolor{black}{##2}\vspace{0.25em}
      },
    },
    joint/.style n args={3}{
      draw = foregroundlib,
      composite={##1}{$|\freeVariables{}|$ = ##2, $|\virtualVariables{}|$ = ##3},
    },
    group/.style n args = {2}{block,
      node contents = {Group: ##1\vspace{##2}},
      draw = foregroundlib, 
      fill = none,
      yshift=0.5*##2+8.5em,
    },
    emptyjoint/.style n args={1}{
      draw = foregroundlib,
      composite={##1}{\vspace{1em}},
    },
    filler/.style = {block,
      text width=2em,
      inner ysep=-0.25em,
      node contents = {$\vdots$\vspace{1.em}},
      draw = foregroundlib, fill = background,
    },
    highlight/.style = {
      draw = darkgreen, fill = lightgreen,
    },
    cross/.style 2 args={
      overlay,
      inner sep=0pt,
      minimum height=##1,
      minimum width=##2,
      yshift=-0.5*##1,
      append after command={
        \pgfextra{
          \draw[black, very thick, -] (\tikzlastnode.south west) -- (\tikzlastnode.north east);
          \draw[black, very thick, -] (\tikzlastnode.north west) -- (\tikzlastnode.south east);
        }
      }
    }
  }
}

%% file: cmds/general.tex
\newcommand{\jointcounter}{n}
\newcommand{\timecounter}{k}
\newcommand{\ntime}{N}
\newcommand{\PredictionHorizon}{\ntime{}}

\NewDocumentCommand{\eqsubscript}{O{}}{\epsilon}
\NewDocumentCommand{\neqsubscript}{O{}}{\xi}

\NewDocumentCommand{\myVector}{ 
    m 
    }{%
    \boldsymbol{#1}
}%
\NewDocumentCommand{\myMatrix}{ 
    m 
    }{%
    \boldsymbol{#1}
}%

\newcommand{\myExpression}[1]{#1}
\newcommand{\myEvaluated}[1]{#1}

\NewDocumentCommand{\myOptions}{ 
    O{} 
    m
}{%
    \IfSubStr{#1}{v}{\myVector}{}
    {
        \IfSubStr{#1}{h}{\hat}{}
        {
            \IfSubStr{#1}{e}
                {\myExpression}
                {\myEvaluated}
            {
                #2
            }
        }
    }
}%

\def\clap#1{\hbox to 0pt{\hss#1\hss}}
\RenewDocumentCommand{\dot}{m}{%
    \overset{\mathrm{\raisebox{-0.15ex}{\clap{$\displaystyle\hspace{0.1em}.$}}}}{#1}
}
\RenewDocumentCommand{\ddot}{m}{%
    \overset{\mathrm{\raisebox{-0.15ex}{\clap{$\displaystyle\hspace{0.1em}.\hspace{-0.08em}.$}}}}{#1}
}
\RenewDocumentCommand{\dddot}{m}{%
    \overset{\mathrm{\raisebox{-0.15ex}{\clap{$\displaystyle\hspace{0.1em}.\hspace{-0.1em}.\hspace{-0.1em}.$}}}}{#1}
}
\RenewDocumentCommand{\ddddot}{m}{%
    \overset{\mathrm{\raisebox{-0.15ex}{\clap{$\displaystyle\hspace{0.1em}.\hspace{-0.12em}.\hspace{-0.12em}.\hspace{-0.12em}.$}}}}{#1}
}
\NewDocumentCommand{\dddddot}{m}{%
    \overset{\mathrm{\raisebox{-0.15ex}{\clap{$\displaystyle\hspace{0.1em}.\hspace{-0.14em}.\hspace{-0.14em}.\hspace{-0.14em}.$}}}}{#1}
}

\NewDocumentCommand{\id}{ 
    O{None} 
    }{%
    \IfEqCase{#1}{%
        {None}{\myMatrix{I}}%
    }[%
        \myMatrix{I}^{#1}
    ]%
}%
\NewDocumentCommand{\zero}{ O{None} O{None}}{%
    \IfEqCase{#2}{%
        {None}{
            \IfEqCase{#1}{%
                {None}{
                    \myMatrix{0}
                }%
            }[%
                \myMatrix{0}^{#1}
            ]%
        }%
    }[%
        \myMatrix{0}^{#1 \times #2}
    ]%
}%
\NewDocumentCommand{\ones}{ O{None} O{None}}{%
    \IfEqCase{#2}{%
        {None}{
            \IfEqCase{#1}{%
                {None}{
                    \myMatrix{1}
                }%
            }[%
                \myMatrix{1}^{#1 \times #1}
            ]%
        }%
    }[%
        \myMatrix{1}^{#1 \times #2}
    ]%
}%

%% file: cmds/transforms.tex
\NewDocumentCommand{\rotation}{ 
    O{None} 
    O{None} 
    }{%
    \IfEq{#1}{None}{%
        \IfEq{#2}{None}{%
            \myMatrix{R}
        }{%
            \GenericError{ }{Error: Either both arguments are None of none is}{}{}
        }%
    }{%
        \IfEq{#2}{None}{%
            \GenericError{ }{Error: Either both arguments are None of none is}{}{}
        }{%
            _{\mathtt{#1}}\myMatrix{R}_{\mathtt{#2}}
        }%
    }%
}%

\NewDocumentCommand{\transform}{ 
    O{None} 
    O{None} 
    }{%
    \IfEq{#1}{None}{%
        \IfEq{#2}{None}{%
            \myMatrix{T}
        }{%
            \GenericError{ }{Error: Either both arguments are None of none is}{}{}
        }%
    }{%
        \IfEq{#2}{None}{%
            \GenericError{ }{Error: Either both arguments are None of none is}{}{}
        }{%
            _{\mathtt{#1}}\myMatrix{T}_{\mathtt{#2}}
        }%
    }%
}%

%% file: cmds/world.tex
\newcommand{\links}{\mathcal{L}}

\newcommand{\joints}{\mathcal{J}}

\newcommand{\symbols}{\mathcal{S}}

\newcommand{\freeVariables}{\mathcal{F}}
\newcommand{\virtualVariables}{\mathcal{V}}
\newcommand{\worldConstraints}{\mathcal{C}}

\NewDocumentCommand{\pointf}{ 
    O{} 
    O{} 
}{%
    \IfValueTF{#1}{%
        \IfValueTF{#2}{%
            {_{#1}\myVector{p}_{#2}}%
        }{%
            \todo{Error: Only parent specified}%
        }%
    }{%
        \IfValueTF{#2}{%
            \todo{Error: Only child specified}%
        }{%
            \myVector{p}%
        }%
    }%
}

\NewDocumentCommand{\vectorf}{ 
    O{} 
    O{} 
}{%
    \IfValueTF{#1}{%
        \IfValueTF{#2}{%
            {_{#1}\myVector{v}_{#2}}%
        }{%
            \todo{Error: Only parent specified}%
        }%
    }{%
        \IfValueTF{#2}{%
            \todo{Error: Only child specified}%
        }{%
            \myVector{v}%
        }%
    }%
}

\NewDocumentCommand{\jointf}{ 
    O{} 
    O{} 
}{%
    \IfValueTF{#1}{%
        \IfValueTF{#2}{%
            {_{#1}\myMatrix{T}_{#2}}%
        }{%
            \todo{Error: Only parent specified}%
        }%
    }{%
        \IfValueTF{#2}{%
            \todo{Error: Only child specified}%
        }{%
            \myMatrix{T}%
        }%
    }%
}

\NewDocumentCommand{\q}{O{}}{\myOptions[#1]{q}}
\NewDocumentCommand{\qd}{O{}}{\dot{\q[#1]{}}}
\NewDocumentCommand{\qdd}{O{}}{\ddot{\q[#1]{}}}
\NewDocumentCommand{\qddd}{O{}}{\dddot{\q[#1]{}}}
\NewDocumentCommand{\qdddd}{O{}}{\ddddot{\q[#1]{}}}
\NewDocumentCommand{\qddddd}{O{}}{\dddddot{\q[#1]{}}}
\NewDocumentCommand{\qaux}{O{}}{\q[#1]_{\mathrm{aux}}}

%% file: cmds/monitors.tex
\NewDocumentCommand{\taskInactive}{ }{%
    \begin{tikzpicture}[bt={5pt}{7pt},]
         \node[lcfsmNotStarted=,rounded corners = 1pt] {};
    \end{tikzpicture}%
}%
\NewDocumentCommand{\taskActive}{ }{%
    \begin{tikzpicture}[bt={5pt}{7pt},]
        \node[lcfsmRunning=,rounded corners = 1pt] {};
    \end{tikzpicture}%
}%
\NewDocumentCommand{\taskOnHold}{ }{%
    \begin{tikzpicture}[bt={5pt}{7pt},]
        \node[lcfsmPaused=,rounded corners = 1pt] {};
    \end{tikzpicture}%
}%
\NewDocumentCommand{\taskDone}{ }{%
    \begin{tikzpicture}[bt={5pt}{7pt},]
        \node[lcfsmStopped=,rounded corners = 1pt] {};
    \end{tikzpicture}%
}%

%% file: cmds/qp.tex
\NewDocumentCommand{\Jacobian}{ 
    O{} 
    O{} 
    O{} 
    O{} 
    }{%
    \metaJacobian[#1][#2][#3][#4][\myMatrix{J}]{}
}%
\NewDocumentCommand{\Gradient}{ 
    O{} 
    O{} 
    O{} 
    O{} 
    }{%
    \metaJacobian[#1][#2][#3][#4][\nabla]{}
}%
\NewDocumentCommand{\metaJacobian}{ 
    O{} 
    O{} 
    O{} 
    O{} 
    O{} 
    }{%
    \IfEqCase{#2}{
        {2}{
            \dot
            }
        {3}{
            \ddot
        }
        {}{}
    }[\overset{#2}]
    {
        #5
    }
    \IfEq{#3}{}{}{
        \IfEq{#4}{}{
            _{#3}
        }{
            _{#3, #4}
        }
    }
    \IfEq{#1}{}{}{(#1)}
}%

\newcommand{\dt}{\Delta t}

\NewDocumentCommand{\xdot}{O{v}}{\myOptions[#1]{x}}
\NewDocumentCommand{\udot}{O{v}}{\myOptions[#1]{u}}

\NewDocumentCommand{\xdotqd}{O{v}}{\xdot[#1]{}_{\qd{}}}

\NewDocumentCommand{\xdotqdd}{O{v}}{\xdot[#1]{}_{\qdd{}}}

\NewDocumentCommand{\xdotqddd}{O{v}}{\xdot[#1]{}_{\qddd{}}}

\NewDocumentCommand{\xdotEqConstr}{O{v}}{\xdot[#1]{}_{=}}
\NewDocumentCommand{\xdotNeqConstr}{O{v}}{\xdot[#1]{}_{\neq}}
\NewDocumentCommand{\xdotVelConstr}{O{v}}{\xdot[#1]{}_{\dot{\neq}}}
\NewDocumentCommand{\xdotAccConstr}{O{v}}{\xdot[#1]{}_{\ddot{\neq}}}
\NewDocumentCommand{\xdotJerkConstr}{O{v}}{\xdot[#1]{}_{\dddot{\neq}}}

\input{cmds/qp/task_functions.tex}

\input{cmds/qp/bounds.tex}

\input{cmds/qp/weights.tex}

\input{cmds/qp/state_model.tex}

\input{cmds/qp/pos_limits.tex}

%% file: cmds/qp/task_functions.tex
\NewDocumentCommand{\refVel}{O{} O{}}{\myOptions[#1]{v}_{#2\mathrm{ref}}}
\NewDocumentCommand{\refVelEqP}{O{}}{\refVel[#1][\eqsubscript{}, p, ]}
\NewDocumentCommand{\refVelNeqP}{O{}}{\refVel[#1][\neqsubscript{}, p, ]}
\NewDocumentCommand{\refVelEqV}{O{}}{\refVel[#1][\eqsubscript{}, v, ]}
\NewDocumentCommand{\refVelNeqV}{O{}}{\refVel[#1][\neqsubscript{}, v, ]}
\NewDocumentCommand{\refVelNeqVel}{O{}}{\refVel[#1]{}_{\dot{\neq}}}
\NewDocumentCommand{\refVelNeqAcc}{O{}}{\refVel[#1]{}_{\ddot{\neq}}}
\NewDocumentCommand{\refVelNeqJerk}{O{}}{\refVel[#1]{}_{\dddot{\neq}}}

\NewDocumentCommand{\slack}{O{}}{\myOptions[#1]{\epsilon}}
\NewDocumentCommand{\slackTwo}{O{}}{\myOptions[#1]{\xi}}
\NewDocumentCommand{\slackEq}{O{}}{\slack[#1]}
\NewDocumentCommand{\slackEqP}{O{}}{\slack[#1]}
\NewDocumentCommand{\slackEqV}{O{} O{\timecounter}}{\slack[#1]_{\eqsubscript{}, v, #2}}
\NewDocumentCommand{\slackNeq}{O{}}{\slackTwo[#1]}
\NewDocumentCommand{\slackNeqP}{O{}}{\slackTwo[#1]}
\NewDocumentCommand{\slackNeqV}{O{} O{\timecounter}}{\slackTwo[#1]_{\neqsubscript{}, v, #2}}

\NewDocumentCommand{\taskF}{O{}}{\myOptions[#1]{f}}

\NewDocumentCommand{\taskFEq}{O{}}{\taskF[#1]{}_{\eqsubscript{}}}
\NewDocumentCommand{\taskFEqP}{O{}}{\taskF[#1]_{\eqsubscript{}}}
\NewDocumentCommand{\taskFEqV}{O{}}{\taskF[#1]_{\eqsubscript{}, v}}

\NewDocumentCommand{\taskFNeq}{O{}}{\taskF[#1]{}_{\neqsubscript{}}}
\NewDocumentCommand{\taskFNeqP}{O{}}{\taskF[#1]_{\neqsubscript{}}}
\NewDocumentCommand{\taskFNeqV}{O{}}{\taskF[#1]_{\neqsubscript{}, v}}

\NewDocumentCommand{\taskFNeqVel}{O{}}{\taskF[#1]{}_{\dot{\neqsubscript{}}}}

\NewDocumentCommand{\taskFNeqAcc}{O{}}{\taskF[#1]{}_{\ddot{\neqsubscript{}}}}

\NewDocumentCommand{\taskFNeqJerk}{O{}}{\taskF[#1]{}_{\dddot{\neqsubscript{}}}}

%% file: cmds/qp/bounds.tex
\NewDocumentCommand{\bound}{ 
    O{} 
}{%
    \myOptions[#1]{b}
}%

\NewDocumentCommand{\bE}{ 
    O{} 
}{%
    \bound[#1]{}_{\slack{}}
}%
\NewDocumentCommand{\bEP}{ 
    O{} 
}{%
    \bE[#1]_{p}
}%
\NewDocumentCommand{\bEV}{ 
    O{} 
}{%
    \bE[#1]_{v}
}%
\NewDocumentCommand{\bEclamped}{ 
    O{} 
}{%
    \bE[#1]^{*}
}%
\NewDocumentCommand{\bEVclamped}{
    O{} 
}{%
    \bEV[#1]^{*}
}%

\NewDocumentCommand{\lb}{ 
    O{} 
}{%
    \myOptions[#1]{l}
}

\NewDocumentCommand{\lbA}{ 
    O{} 
}{%
    \myOptions[#1]{l}
}

\NewDocumentCommand{\lbASlack}{ 
    O{} 
}{%
    \lbA[#1]_{\slackTwo{}}
}
\NewDocumentCommand{\lbAP}{ 
    O{} 
}{%
    \lbA[#1]_{p}
}
\NewDocumentCommand{\lbAV}{ 
    O{} 
}{%
    \lbA[#1]_{v}
}
\NewDocumentCommand{\lbAVclamped}{
    O{} 
}{%
    \lbAV[#1]^{*}
}

\NewDocumentCommand{\lbAclamped}{ 
    O{} 
}{%
    \lbA[#1]^{*}
}

\NewDocumentCommand{\ub}{ 
    O{} 
}{%
    \myOptions[#1]{u}
}%

\NewDocumentCommand{\ubA}{ 
    O{} 
}{%
    \myOptions[#1]{u}
}%

\NewDocumentCommand{\ubASlack}{ 
    O{} 
}{%
    \ubA[#1]_{\slackTwo{}}
}%

\NewDocumentCommand{\ubAP}{ 
    O{} 
}{%
    \ubA[#1]_{p}
}
\NewDocumentCommand{\ubAV}{ 
    O{} 
}{%
    \ubA[#1]_{v}
}
\NewDocumentCommand{\ubAVclamped}{
    O{} 
}{%
    \ubAV[#1]^{*}
}

\NewDocumentCommand{\ubAclamped}{ 
    O{} 
}{%
    \ubA[#1]^{*}
}

\NewDocumentCommand{\lbANeq}{ 
    O{} 
}{%
    \lbA[#1]{}_{\neqModel{}}
}

\NewDocumentCommand{\ubANeq}{ 
    O{} 
}{%
    \ubA[#1]{}_{\neqModel{}}
}%

\NewDocumentCommand{\lbANeqVel}{ 
    O{} 
}{%
    \lbA[#1]{}_{\neqVelModel{}}
}
\NewDocumentCommand{\ubANeqVel}{ 
    O{} 
}{%
    \ubA[#1]{}_{\neqVelModel{}}
}%

\NewDocumentCommand{\lbANeqAcc}{ 
    O{} 
}{%
    \lbA[#1]{}_{\neqAccModel{}}
}

\NewDocumentCommand{\ubANeqAcc}{ 
    O{} 
}{%
    \ubA[#1]{}_{\neqAccModel{}}
}%

\NewDocumentCommand{\lbANeqJerk}{ 
    O{} 
}{%
    \lbA[#1]{}_{\neqJerkModel{}}
}
\NewDocumentCommand{\ubANeqJerk}{ 
    O{} 
}{%
    \ubA[#1]{}_{\neqJerkModel{}}
}%

\NewDocumentCommand{\lbAPos}{ 
    O{} 
}{%
    \lbA[#1]{}_{\q{}}
}
\NewDocumentCommand{\ubAPos}{ 
    O{} 
}{%
    \ubA[#1]{}_{\q{}}
}%

\NewDocumentCommand{\lbAVel}{ 
    O{} 
    O{} 
}{%
    \lbA[#1]{}_{\qd{} #2}
}
\NewDocumentCommand{\ubAVel}{ 
    O{} 
    O{} 
}{%
    \ubA[#1]{}_{\qd{} #2}
}%

\NewDocumentCommand{\lbAAcc}{ 
    O{} 
}{%
    \lbA[#1]{}_{\qdd{}}
}
\NewDocumentCommand{\ubAAcc}{ 
    O{} 
}{%
    \ubA[#1]{}_{\qdd{}}
}%

\NewDocumentCommand{\lbAJerk}{ 
    O{} 
}{%
    \lbA[#1]{}_{\qddd{}}
}
\NewDocumentCommand{\ubAJerk}{ 
    O{} 
}{%
    \ubA[#1]{}_{\qddd{}}
}%

\NewDocumentCommand{\lbSlackEq}{ 
    O{} 
}{%
    \lbA[#1]{}_{=}
}%
\NewDocumentCommand{\ubSlackEq}{ 
    O{} %
}{%
    \ubA[#1]{}_{=}
}%
\NewDocumentCommand{\lbSlackNeq}{ 
    O{} 
}{%
    \lbA[#1]{}_{\neq}
}%
\NewDocumentCommand{\ubSlackNeq}{ 
    O{} %
}{%
    \ubA[#1]{}_{\neq}
}%

\NewDocumentCommand{\lbSlackVel}{ 
    O{} 
}{%
    \lbA[#1]{}_{\dot{\neq}}
}%
\NewDocumentCommand{\ubSlackVel}{ 
    O{} %
}{%
    \ubA[#1]{}_{\dot{\neq}}
}%

\NewDocumentCommand{\lbSlackAcc}{ 
    O{} 
}{%
    \lbA[#1]{}_{\ddot{\neq}}
}%
\NewDocumentCommand{\ubSlackAcc}{ 
    O{} %
}{%
    \ubA[#1]{}_{\ddot{\neq}}
}%

\NewDocumentCommand{\lbSlackJerk}{ 
    O{} 
}{%
    \lbA[#1]{}_{\dddot{\neq}}
}%
\NewDocumentCommand{\ubSlackJerk}{ 
    O{} %
}{%
    \ubA[#1]{}_{\dddot{\neq}}
}%

\NewDocumentCommand{\bLink}{ 
    O{None} 
    }{%
    \bound[v]{}
    \IfEqCase{#1}{%
        {None}{
            _{\stateModel{}}{}
        }%
        {1}{
            _{\stateModel[1]{}}
        }%
        {2}{
            _{\stateModel[2]{}}
        }
    }[%
    ]%
}%

\NewDocumentCommand{\defaultVel}{ 
O{}
O{} 
}{%
    \qd[#1]{}_{
        \IfSubStr{#1}{v}{}{\jointcounter{}}
    }^{\texttt{max}}
}%
\NewDocumentCommand{\defaultAcc}{ 
O{}
O{} 
}{%
    \qdd[#1]{}_{
        \IfSubStr{#1}{v}{}{\jointcounter{}}
    }^{\texttt{max}}
}%
\NewDocumentCommand{\defaultJerk}{ 
O{}
O{} 
}{%
    \qddd[#1]{}_{
        \IfSubStr{#1}{v}{}{\jointcounter{}}
    }^{\texttt{max}}
}%

%% file: cmds/qp/weights.tex
\NewDocumentCommand{\weights}{O{}}{\myOptions[#1]{w}}

\NewDocumentCommand{\wqd}{O{v} O{}}{\weights[#1]{}_{\qd[]{}#2}}
\NewDocumentCommand{\wqdnoT}{O{v} O{}}{\weights[#1]{}_{\qd[]{}#2}}
\NewDocumentCommand{\wqdDefault}{O{v}}{\defaultWeight[#1][\qd[]{}]{}}
\NewDocumentCommand{\wqdd}{O{v} O{}}{\weights[#1]{}_{\qdd[]{}#2}}
\NewDocumentCommand{\wqddd}{O{v} O{}}{\weights[#1]{}_{\qddd[]{}#2}}

\NewDocumentCommand{\wEq}{O{v} O{}}{\weights[#1]{}_{\eqsubscript{}#2}}
\NewDocumentCommand{\wEqP}{O{v} O{}}{\weights[#1]{}_{\eqsubscript{}#2}}
\NewDocumentCommand{\wEqDefault}{O{v}}{\defaultWeight[#1][\eqsubscript{}]{}}
\NewDocumentCommand{\wNeq}{O{v} O{}}{\weights[#1]{}_{\neq#2}}
\NewDocumentCommand{\wNeqP}{O{v} O{}}{\weights[#1]{}_{\neqsubscript{}#2}}

\NewDocumentCommand{\wEqV}{O{v} O{}}{\weights[#1]{}_{\eqsubscript{}, v#2}}
\NewDocumentCommand{\wNeqV}{O{v} O{}}{\weights[#1]{}_{\neqsubscript{}, v#2}}
\NewDocumentCommand{\wNeqVel}{O{v} O{}}{\weights[#1]{}_{\dot{\neq}#2}}
\NewDocumentCommand{\wNeqVelDefault}{O{v}}{\defaultWeight[#1][\dot{\neq}]{}}
\NewDocumentCommand{\wNeqAcc}{O{v} O{}}{\weights[#1]{}_{\ddot{\neq}#2}}
\NewDocumentCommand{\wNeqJerk}{O{v} O{}}{\weights[#1]{}_{\dddot{\neq}#2}}

\NewDocumentCommand{\Weights}{O{v}}{\myOptions[#1]{W}}

\NewDocumentCommand{\Wq}{O{v}}{\Weights[#1]{}_{\q[]{}}}
\NewDocumentCommand{\Wqd}{O{v}}{\Weights[#1]{}_{\qd[]{}}}
\NewDocumentCommand{\Wqdd}{O{v}}{\Weights[#1]{}_{\qdd[]{}}}
\NewDocumentCommand{\Wqddd}{O{v}}{\Weights[#1]{}_{\qddd[]{}}}

\NewDocumentCommand{\WEq}{O{v}}{\Weights[#1]{}_{\eqsubscript{}}}
\NewDocumentCommand{\WEqP}{O{v}}{\Weights[#1]{}_{\eqsubscript{}}}
\NewDocumentCommand{\WEqV}{O{v}}{\Weights[#1]{}_{\eqsubscript{} v}}
\NewDocumentCommand{\WNeq}{O{v}}{\Weights[#1]{}_{\neqsubscript{}}}
\NewDocumentCommand{\WNeqP}{O{v}}{\Weights[#1]{}_{\neqsubscript{}}}
\NewDocumentCommand{\WNeqV}{O{v}}{\Weights[#1]{}_{\neqsubscript{} v}}

\NewDocumentCommand{\WEqVel}{O{v}}{\WEq[#1]_{v}}
\NewDocumentCommand{\WNeqVel}{O{v}}{\WNeq[#1]_{v}}

\NewDocumentCommand{\WNeqAcc}{O{v}}{\Weights[#1]{}_{\ddot{\neq}}}
\NewDocumentCommand{\WNeqJerk}{O{v}}{\Weights[#1]{}_{\dddot{\neq}}}

\NewDocumentCommand{\defaultWeight}{ 
    O{None} 
    O{None} 
}{%
    \weights[#1]{}^{\texttt{def}}_{
        #2_{
            \IfSubStr{#1}{v}{}{\jointcounter{}}
        }
    }
}%

%% file: cmds/qp/state_model.tex
\NewDocumentCommand{\stateModel}{ 
    O{None} 
    }{%
    \IfEqCase{#1}{%
        {None}{
            \myMatrix{M}
        }%
        {1}{
            \myMatrix{M}_{\qd{}}
        }%
        {2}{
            \myMatrix{M}_{\qdd{}}
        }
    }[%
    ]%
}%
\NewDocumentCommand{\eqModel}{ 
    O{None} 
    }{%
    \IfEqCase{#1}{%
        {None}{
            \myMatrix{E}
        }%
        {1}{
            \myMatrix{E}_{\qd{}}
        }%
        {2}{
            \myMatrix{E}_{\qdd{}}
        }
    }[%
    ]%
}%
\NewDocumentCommand{\neqModel}{ 
    O{None} 
    }{%
    \IfEqCase{#1}{%
        {None}{
            \myMatrix{N}
        }%
        {1}{
            \myMatrix{N}_{\qd{}}
        }
        {2}{
            \myMatrix{N}_{\qdd{}}
        }
    }[%
    ]%
}%
\NewDocumentCommand{\neqVelModel}{ 
    O{None} 
    }{%
    \IfEqCase{#1}{%
        {None}{
            \myMatrix{\dot{N}}
        }%
        {1}{
            \myMatrix{\dot{N}}_{\qd{}}
        }
        {2}{
            \myMatrix{\dot{N}}_{\qdd{}}
        }
    }[%
    ]%
}%
\NewDocumentCommand{\neqAccModel}{ 
    O{None} 
    }{%
    \IfEqCase{#1}{%
        {None}{
            \myMatrix{\ddot{N}}
        }%
        {1}{
            \myMatrix{\ddot{N}}_{\qd{}}
        }
        {2}{
            \myMatrix{\ddot{N}}_{\qdd{}}
        }
    }[%
    ]%
}%
\NewDocumentCommand{\neqJerkModel}{ 
    O{None} 
    }{%
    \IfEqCase{#1}{%
        {None}{
            \myMatrix{\dddot{N}}
        }%
        {1}{
            \myMatrix{\dddot{N}}_{\qd{}}
        }
        {2}{
            \myMatrix{\dddot{N}}_{\qdd{}}
        }
    }[%
    ]%
}%

%% file: cmds/qp/pos_limits.tex
\NewDocumentCommand{\velProfile}{O{None} O{None} O{None}}{\Profile[#1][#2][#3][v]{}}
\NewDocumentCommand{\accProfile}{O{None} O{None} O{None}}{\Profile[#1][#2][#3][a]{}}
\NewDocumentCommand{\jerkProfile}{O{None} O{None} O{None}}{\Profile[#1][#2][#3][j]{}}

\NewDocumentCommand{\Profile}{ 
    O{None} 
    O{None} 
    O{None} 
    O{v} 
}{%
    \IfEqCase{#1}{%
    {implicit}{
        \mathbf{#4}^{\texttt{imp}}
    }%
    {shifted}{
        \mathbf{#4}^{
        \IfEq{#3}{None}{
        }{
            #3
        }
        }
    }%
    {recovery}{
        \mathbf{#4}^{\texttt{rcv}}
    }%
    {ref}{
        \mathbf{#4}^{\texttt{ref}}
    }%
    {max}{
        \mathbf{#4}^{\texttt{max}}
    }%
    {asap}{
        \mathbf{#4}^{\texttt{asap}}
    }%
    {result}{
        \mathbf{#4}^{\texttt{result}}
    }%
    {default}{
        \mathbf{#4}^{\texttt{def}}
    }
    }[%
    ]%
    \IfEq{#2}{None}{
    }{
        _{#2}
    }
}%

%% file: acronyms.tex
\acrodef{dof}[DoF]{Degree of Freedom}
\acrodefplural{dof}[DoFs]{Degrees of Freedom}

\acrodef{fk}[FK]{forward kinematics}

\acrodef{fsm}[FSM]{finite state machine}
\acrodef{bt}[BT]{Behavior Tree}
\acrodef{msc}[MSC]{Motion Statechart}

\acrodef{cram}[CRAM]{Cognitive Robot Abstract Machine}
\acrodef{pddl}[PDDL]{Planning Domain Definition Language}

\acrodef{ompl}[OMPL]{Open Motion Planning Library}

\acrodef{hfsm}[HFSM]{Hierarchical Finite State Machine}

\acrodef{ik}[IK]{Inverse Kinematics}

\acrodef{json}[JSON]{JavaScript Object Notation}

\acrodef{vla}[VLA]{Vision-Language-Action model}

\acrodef{ros}[ROS]{Robot Operating System}

\acrodef{urdf}[URDF]{Unified Robot Description Format}

\acrodef{sdf}[SDF]{Simulation Description Format}

\acrodef{mpc}[MPC]{Model Predictive Control}

\acrodef{lmpc}[lMPC]{linear Model Predictive Control}

\acrodef{qp}[QP]{Quadratic Program}

\acrodef{etasl}[eTaSL/eTC]{expressiongraph-based Task Specification Language / expressiongraph-based Task Controller}